\documentclass[10pt,twocolumn,letterpaper]{article}

\usepackage[pagenumbers]{cvpr} 

\usepackage{multirow}
\usepackage{colortbl}
\usepackage[table,xcdraw,x11names]{xcolor}
\usepackage{algpseudocode}
\definecolor{commgreen}{HTML}{009933}
\usepackage[linesnumbered,ruled,vlined]{algorithm2e}
\usepackage[accsupp]{axessibility}  

\newcommand{\PAR}[1]{\vskip3pt \noindent{\bf #1~}}
\SetCommentSty{licommfont}
\newcommand{\samplfull}{Pseudocode for RSD}
\definecolor{cvprblue}{rgb}{0.21,0.49,0.74}
\usepackage[pagebackref,breaklinks,colorlinks,allcolors=cvprblue]{hyperref}
\newcommand{\greencell}{\cellcolor{Green2!25}}
\newcommand{\orangecell}{\cellcolor{Orange1!40}}
\newcommand{\redcell}{\cellcolor{IndianRed1!40}}
\def\paperID{2445} 
\def\confName{CVPR}
\def\confYear{2025}

\title{LightLoc: Learning Outdoor LiDAR Localization at Light Speed}

\begin{document}
\author{Wen Li$^{1,2}$\footnotemark[1]\hspace{4mm}
Chen Liu$^{1,2}$\footnotemark[1]\hspace{4mm}
Shangshu Yu$^{3}$\footnotemark[2]\hspace{4mm}
Dunqiang Liu$^{1,2}$\hspace{4mm}
Yin Zhou$^4$\hspace{4mm}\\
Siqi Shen$^{1,2}$\hspace{4mm}
Chenglu Wen$^{1,2}$\hspace{4mm}
Cheng Wang$^{1,2}$\footnotemark[2]\\
$^1$Fujian Key Laboratory of Sensing and Computing for Smart Cities,  Xiamen University\\
$^2$Key Laboratory of Multimedia Trusted Perception and Efficient Computing, \\
Ministry of Education of China, Xiamen University\\
$^3$Nanyang Technological University
$^4$GAC R\&D Center\\
}
\maketitle

\footnotetext[1]{Equal contribution.}
\footnotetext[2]{Corresponding author.} 
\begin{abstract}
Scene coordinate regression achieves impressive results in outdoor LiDAR localization but requires days of training. 
Since training needs to be repeated for each new scene, long training times make these impractical for applications requiring time-sensitive system upgrades, such as autonomous driving, drones, robotics, etc.
We identify large coverage areas and vast amounts of data in large-scale outdoor scenes as key challenges that limit fast training.
In this paper, we propose LightLoc, the first method capable of efficiently learning localization in a new scene at light speed.
Beyond freezing the scene-agnostic feature backbone and training only the scene-specific prediction heads, we introduce two novel techniques to address these challenges. 
First, we introduce sample classification guidance to assist regression learning, reducing ambiguity from similar samples and improving training efficiency.
Second, we propose redundant sample downsampling to remove well-learned frames during training, reducing training time without compromising accuracy.
In addition, the fast training and confidence estimation characteristics of sample classification enable its integration into SLAM, effectively eliminating error accumulation. 
Extensive experiments on large-scale outdoor datasets demonstrate that LightLoc achieves state-of-the-art performance with just \textbf{1 hour} of training—$\textbf{50}\times$ faster than existing methods. 
Our Code is available at \url{https://github.com/liw95/LightLoc}.
\end{abstract} 
\section{Introduction}
\label{sec:intro}

LiDAR localization aims to estimate the 6-DoF pose of sensors, which is a fundamental component of many applications, \emph{e.g.,} autonomous driving~\cite{Yin_2024_IJCV} and robotics~\cite{Zhang_2024_AAAI}.

\begin{figure}[t]
  \centering
  \includegraphics[width=1\columnwidth]{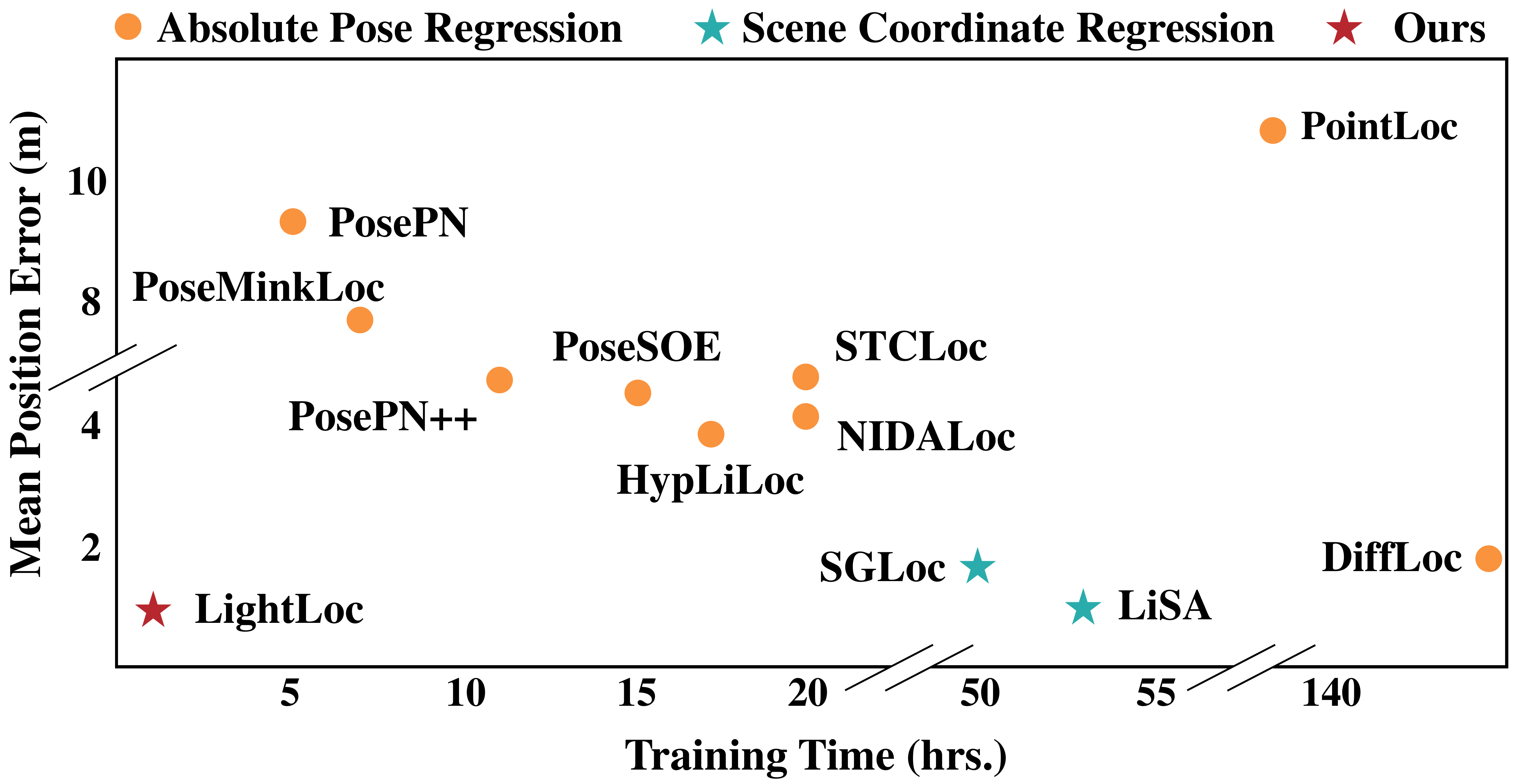}
   \vspace{-0.6cm}
   \caption{
   \textbf{LiDAR localization performance vs. training time}. 
   The figure shows the mean position error and training time of several regression-based methods on the QEOxford~\cite{Dan_2020_ICRA, Li_2023_CVPR} dataset. The proposed LightLoc achieves state-of-the-art performance with significantly reduced training time.
   }
   \label{fig:1}
\end{figure}

Contemporary state-of-the-art LiDAR-based localization
methods rely on explicit map usage, matching query points with a pre-built 3D map~\cite{Xia_2021_CVPR, Xia_2023_ICCV, Wang_2024_AAAI}. 
However, these approaches often demand costly 3D map storage and high communication overhead.
Therefore, regression-based methods~\cite{Brahmbhatt_2018_CVPR, Chen_2024_CVPR, Li_2024_CVPR, Brachmann_2023_CVPR, Yang_2024_CVPR} have been proposed to address these limitations by memorizing the specific scene within a network. 
During inference, these methods eliminate the need for pre-built map matching, thus reducing both storage and communication requirements.  
Based on their regression objectives, these methods can be further categorized into Absolute Pose Regression (APR)~\cite{Brahmbhatt_2018_CVPR, Chen_2024_CVPR, Li_2024_CVPR} and 
Scene Coordinate Regression (SCR)~\cite{Brachmann_2023_CVPR, Li_2023_CVPR, Yang_2024_CVPR}. 
Specifically, APR directly regresses the poses, while SCR predicts correspondences and uses RANASC~\cite{Fischler_1981_Commun} to solve the poses. 
Among them, 
SCR explicitly leverages geometric information, embedding scene geometry into its parameters~\cite{Li_2023_CVPR}, resulting in higher localization accuracy. 

Although SCR performs well, 
due to training needs to be repeated for each new scene, long training times make it impractical for applications.
For example, LiSA~\cite{Yang_2024_CVPR}, a state-of-the-art method achieving 0.95m accuracy, requires approximately 53 hours of training
, as shown in Fig.~\ref{fig:1}.
Previous efforts~\cite{Brachmann_2023_CVPR, Nguyen_2024_WACV, Wang_2024_CVPR, Chen_2024_CVPR} in camera localization build a buffer on GPUs to store sample features and optimize multiple views simultaneously, accelerating training.
However, applying this 
to large-scale outdoor scenes presents significant challenges.
Specifically, for the autonomous driving dataset~\cite{Dan_2020_ICRA, Li_2023_CVPR}, which covers 2km$^2$ and contains approximately 150K training samples, even sampling sample as 1024 points with 512-dimensional features results in a storage requirement of around 150GB. This makes it difficult to store the features on GPUs.
For large-scale outdoor LiDAR localization, we identify two key challenges in accelerating training: (1) the extensive coverage area, which contains many visually similar regions, complicating the training of regression-based methods, and (2) the vast amount of data, which leads to substantial computational and storage demands. These challenges motivate us to explore methods for accelerating training in large-scale outdoor scenes.

This paper proposes a novel framework, LightLoc, which can efficiently learn localization in a new scene at light speed.
Following previous work~\cite{Brachmann_2023_CVPR, Yu_2022_PR}, we freeze the scene-agnostic feature backbone and train only the scene-specific prediction heads.
Beyond this design, we propose two innovative techniques to accelerate training for large-scale outdoor scenes. 
The first one addresses challenge (1) by building a sample classification task to help the regression training.
We train the classification network for just 5 minutes, and this model is then used to guide the regression network, reducing the ambiguity caused by visually similar regions and enhancing training efficiency.
The second one tackles challenge (2) by a novel redundant sample downsampling technique, which leverages the variance of the median loss to discard well-learned frames during training, without compromising accuracy.
LightLoc achieves state-of-the-art performance with approximately a \textbf{50$\times$} reduction in training time, as shown in Fig.~\ref{fig:1}.
In addition, due to the fast training and confidence estimation characteristics of the proposed sample classification guidance, we introduce it into SLAM~\cite{Zhang_2014_RSS} to eliminate error accumulation.

Our contributions can be summarized as follows:
\begin{itemize}[leftmargin=2em]

\item LightLoc, a LiDAR localization method, can learn a large-scale outdoor scene in just \textbf{1 hour}, 
while the previous state-of-the-art SCR method requires about 2 days of training to achieve comparable accuracy.

\item We introduce sample classification guidance and redundant sample downsampling techniques to address the challenge of large coverage and vast amounts of data in large-scale outdoor LiDAR localization.

\item Leveraging the characteristics of the proposed sample classification guidance, including its 5-minute training time and confidence estimation, we show how it can be integrated into SLAM to mitigate error accumulation.

\end{itemize}
\section{Related Work}
\subsection{Map-based Methods}
Map-based LiDAR localization methods rely on explicit map usage, where query points are matched with a pre-built 3D map to estimate poses. These methods can be further categorized into retrieval-based~\cite{Yu_2021_ISPRS, Xia_2021_CVPR, Xia_2023_ICCV, Wang_2024_AAAI, Komorowski_2021_WACV, Uy_2018_CVPR, Luo_2023_ICCV} and registration-based~\cite{Wang_2019_ICCV, Choy_2019_CVPR, Qin_2022_CVPR, Ao_2023_CVPR, Zhang_2023_CVPR, Jin_2024_CVPR} methods. Retrieval-based methods treat localization as a place recognition problem, searching for frames most similar to the query frame in a pre-built 
database. 
Registration-based methods 
use feature matching to establish correspondences between the query frame and the point cloud. However, both approaches often require costly 3D map storage and high communication overhead.

\subsection{Regression-based Methods}
To solve the limitations of map-based methods, regression-based methods have been proposed and received much attention recently. 
These methods can be further separated into two categories: 
Absolute Pose Regression (APR) and Scene Coordinate Regression (SCR).

\noindent \textbf{Absolute Pose Regression.}
APR directly regresses the global 6-DoF poses of sensors in an end-to-end manner~\cite{Kendall_2015_ICCV, Brahmbhatt_2018_CVPR, Wang_2020_AAAI, Yu_2022_PR, Wang_2023_AAAI, Chen_2024_CVPRB, Chen_2024_CVPR, Chen_2022_ECCV, Shavit_2021_ICCV, Wang_2020_AAAI}. These methods generally follow the same pipeline~\cite{Sattler_2019_CVPR}, where CNNs first learn high-dimensional scene feature representations, and MLP is then used to regress the poses.

PointLoc~\cite{Wang_2022_Sensors} is the pioneer in the investigation of APR for LiDAR Localization.  PointLoc uses PointNet++~\cite{Qi_2017_NIPS} to learn the global feature from the query point cloud, and use the self-attention mechanism~\cite{Vaswani_2017_NIPS} to further refine features~. 
LiDAR-based methods have proven to be more robust to illumination changes than camera-based APR in large-scale outdoor scenarios, and have thus attracted significant attention. The variant~\cite{Yu_2022_PR} investigates different backbones to enhance performance.
HypLiLoc~\cite{Wang_2023_CVPR} fuses multi-modal features in both hyperbolic and Euclidean spaces to further boost performance. These methods take a single-scan point cloud as input, with training times ranging from 5 to 17 hours and achieving an accuracy of around 4 meters.

Some methods~\cite{Yu_2022_TITS, Yu_2023_TITS, Li_2024_CVPR} use multi-frame point clouds as input to leverage temporal constraints for smoothing the predicted poses. DiffLoc~\cite{Li_2024_CVPR}, the state-of-the-art method in APR, formulates pose regression as a conditional generation of poses. DiffLoc achieves 2-meter localization accuracy, but the training time increases significantly, extending to nearly 1 week, as shown in Fig.~\ref{fig:1}.

\begin{figure*}
  \centering
  \includegraphics[width=1\linewidth]{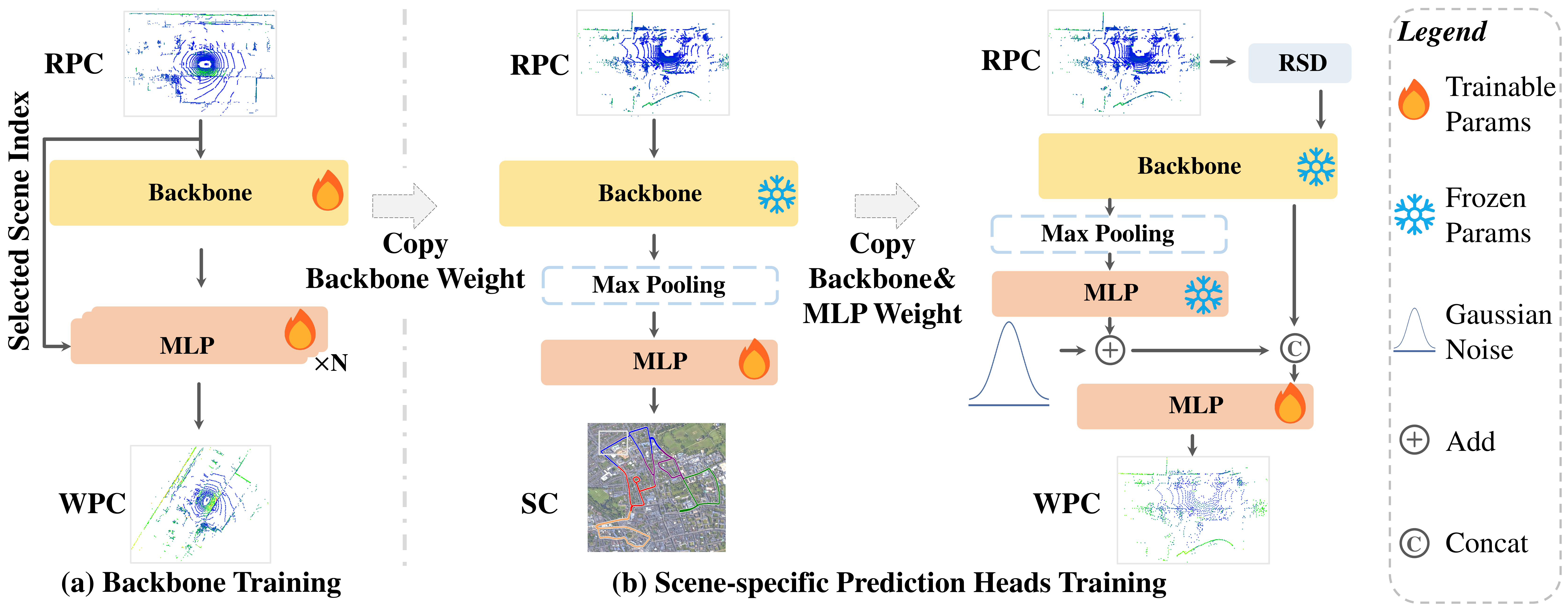}
  \caption{
\textbf{Illustration of the training pipeline for LightLoc}. (a) The backbone is trained with $N$ regression heads for $N$ scenes in parallel to produce a scene-agnostic feature backbone. (b) In new scenes, the backbone parameters are frozen, and only the scene-specific prediction heads are trained. We propose a sample classification guidance (SCG) and a redundant sample downsampling (RSD) technique to accelerate training.
SCG is established by training an MLP head, with the resulting sample probability distribution feature to help SCR learning. RSD is incorporated into the training loop to filter out well-learned samples, enabling high-speed training. \textbf{RPC} and \textbf{WPC} denote the point cloud in raw and world coordinate frames, respectively. \textbf{SC} means sample classification. 
}
  \label{fig:2}
\end{figure*}

\noindent \textbf{Scene Coordinate Regression.}
Unlike APR, SCR~\cite{Li_2023_CVPR, Yang_2024_CVPR} predicts point cloud correspondences between LiDAR and world coordinates, using RANSAC~\cite{Fischler_1981_Commun} to estimate poses.

SGLoc~\cite{Li_2023_CVPR} decouples the pose estimation to point cloud correspondence regression and pose estimation via RANSAC. This decoupling can effectively capture the scene geometry and get a robust localization result, achieving 1.5-meter accuracy in QEOxford~\cite{Dan_2020_ICRA, Li_2023_CVPR}. To further raise performance, LiSA~\cite{Yang_2024_CVPR}, the state-of-the-art method in SCR, introduces semantic awareness into SGLoc and gets about 38$\%$ improvement in position, with 0.95m accuracy.
While achieving higher accuracy than APR, the training time of these methods increases significantly to about 2 days, as shown in Fig.~\ref{fig:1}.

Recently, in camera localization, several works~\cite{Brachmann_2023_CVPR, Wang_2024_CVPR, Nguyen_2024_WACV} are proposed to address the issue of long SCR training time. ACE~\cite{Brachmann_2023_CVPR} formulates SCR as a scene-agnostic feature backbone with a scene-specific prediction head. It stores sample features on GPUs and simultaneously optimizes multiple views during each iteration, thus accelerating training. 
However, when applied to large outdoor scene datasets~\cite{Dan_2020_ICRA, Carlone_2016_IJRR} with 150K training samples, requiring approximately 150GB of storage on GPUs, this becomes a significant challenge.
We propose LightLoc, which incorporates a sample classification guidance and a redundant sample downsampling technique to address the challenges of large-scale outdoor scenes. LightLoc eliminates the need for expensive storage and accelerates training to just 1 hour.
\section{Method}
LiDAR-based SCR demonstrates impressive localization performance. However, they often require up to 2 days of training for pose estimation, due to the challenges posed by large coverage areas and vast amounts of data in large-scale outdoor scenes.
In this paper, we propose LightLoc, which introduces a sample classification guidance (SCG) to help SCR learning, and incorporates a redundant sample downsampling (RSD) to cut well-learned data in training (Sec.~\ref{Sec.3.1}).
Additionally, due to the fast training and confidence estimation characteristics of SCG, we adopt it to SLAM to eliminate error accumulation (Sec.~\ref{Sec.3.2}).

\subsection{LightLoc}\label{Sec.3.1}
We now elaborate LightLoc, as shown in Fig.~\ref{fig:2}, which can be divided into two components. (1) Backbone training, preparing scene-agnostic feature backbone for new scenes. (2) scene-specific prediction heads training, 
by frozen backbone parameters, we utilize the proposed SCG and RSD to achieve training at light speed.

\noindent \textbf{Backbone Training.}
Previous work in LiDAR localization~\cite{Yu_2022_PR} demonstrates that the feature backbone trained on one scene could be transferred to another. 
However, their approach focuses on APR, which extracts a global descriptor for the entire input point cloud. In contrast, SCR requires descriptors that are unique to each position in the point cloud. This motivates us to train a scene-agnostic backbone specifically for SCR.

For the network architecture, we adopt the backbone of SGLoc~\cite{Li_2023_CVPR} with a reduced feature dimension and fewer residual layers to minimize the number of parameters and decrease training time. 
Following ACE~\cite{Brachmann_2023_CVPR}, we train $N$ regression heads in parallel, each corresponding to one of $N$ scenes. This parallel training strategy forces the backbone to learn features that generalize effectively across a diverse array of scenes.
Specifically, we configure the backbone $f$ with $N$ regression heads $h$. 
For each input point cloud ${\cal P}={\{p_i\}}_{i=1}^M$, where $p_i \in \mathbb{R}^{3}$ represents a 3D point in the LiDAR coordinate frame with the LiDAR scanner at the origin, $f$ is used to extract features. 
Based on the scene index $idx$, the corresponding regression head $h_{idx}$ is then selected to predict the point cloud in the world coordinate
$h_{idx}\left(f\left( {\cal P}\right)\right)={\{p_i^{'}\}}_{i=1}^{M'}$, where $M'= M/8$. The training objective is to minimize the $L_1$ distance of the ground truth $p_i^{*} \in \cal P^{*}$ and the prediction $p_i^{'} \in {\cal P}'$ as follows:
\begin{equation}
{\cal L}_{reg}= \frac{\sum_{i=1}^{M^{'}} \Vert {p_{i}^{'} - p_{i}^{*}}\Vert_1} 
{M^{'}}.
\label{eq1}
\end{equation}

We train the backbone $f$ on 18 scenes from nuScenes dataset~\cite{Caesar_2020_CVPR} with 350K frames for 2 days, resulting in 16M of weight that can be used to extract descriptors for new scenes. More details on the backbone architecture and training process can be found in the supplementary.

\noindent \textbf{Scene-specific Prediction Heads Training.}
After obtaining the scene-agnostic feature backbone, 
following previous work~\cite{Yu_2022_PR, Brachmann_2023_CVPR}, we freeze its parameters for new scenes and train only scene-specific prediction heads. However, 
two challenges remain for accelerating SCR training in large-scale outdoor scenes: the extensive coverage area, which complicates regression learning, and the vast amount of data, which imposes a significant computational burden. To address these issues, we propose SCG and RSD techniques.

\textbf{For (1)}, SCG is designed to generate a sample probability distribution feature in just 5 minutes of training. These features are then utilized to assist the SCR's training, drastically reducing the time required to adapt to new scenes. As shown in the left sub-figure of Fig.~\ref{fig:2}~(b), the training objective is formulated as a sample classification task.

Firstly, we define the classification labels. In this paper, we apply the classic K-Means algorithm to distribute the training position into $k_1$ clusters. This method offers two advantages: label generation is cost-free, and clustering on the position rather than the point cloud map ensures a short label generation time. 

Subsequently, we train the classification network. 
For the input point cloud, denoted as $\cal P$, we use the scene-agnostic feature backbone $f$ to extract features $f({\cal P})$. Global max pooling is then applied to obtain global features. Finally, an MLP is used to generate a sample probability distribution. The MLP is trained using cross-entropy loss with a smoothing factor $\epsilon=0.1$ as follows~\cite{Muller_2019_NIPS}:
\begin{equation}
{\cal L}_{cls}= -\sum_{i=1}^{k_1} \left(l^{*}_i\left(1-\epsilon\right) + \frac{\epsilon}{k_1}\right)log{\left(l^{'}_i\right)},
\label{eq2}
\end{equation}

\noindent where $l^{'}_i$ and $l^{*}_i$ denotes the predicted probability and label of class $i$, respectively.


Finally, the normalized sample probability distribution feature is employed to guide regression learning. As shown in the right sub-figure of Fig.~\ref{fig:2}~(b), following GLACE~\cite{Wang_2024_CVPR}, Gaussian noise with a standard deviation of $\sigma=0.1$ is added. After noise addition, the features are normalized back to the unit sphere.

With SCG, we obtain sample probability distribution features. These features are refined with Gaussian noise, normalized to the unit sphere, and then incorporated into the SCR framework. This process effectively guides learning, ensuring fast convergence while preventing overfitting.

\begin{algorithm}[t]
    \SetKwInput{KwInput}{Input}
    \SetKwInput{KwOutput}{Return}
    \DontPrintSemicolon
    \SetAlgoLined
    \SetNoFillComment
    \caption{{\samplfull}.}
    \label{alg:1}
    \SetAlgoVlined
    \KwInput{Training set $\cal T$, downsampling ratio $r_d$, start ratio $r_{st}$, stop ratio $r_{sp}$, training epochs $E$, length of sliding window $S$.}
    First downsampling epochs $E_1\gets\lceil E\times r_{st}\rceil$. \\
    Second downsampling epochs $E_2\gets\lfloor E\times \frac{r_{st}+r_{sp}}{2}\rfloor$.\\
    Stop downsampling epochs $E_s \gets \lfloor E\times r_{sp} \rfloor$. \\
    $\cal T^{'} \gets \cal T$. \\
    \For {$e=0, \ldots, E-1$}{
        $\triangleright$ In parallel on GPUs.\\
		Sample a batch $\cal{B}\sim \cal{T}$. \\
        \For{$({\cal P}_i,{\cal P}_i^{*}) \in \cal{B}$}{
            Compute the median of $L_1$ loss as ${\cal L}_m$. \\
            \If{$e \in \left[ E_j, E_j+S\right)$}
            {\tcp{$j \in \left[1,2 \right]$} 
            Compute the variance $\cal V$ of ${\cal L}_m$ in $S$.
            }
        }
        \If{$e = E_j + S$}{
        Sort ${\cal T}^{'}$ in descending order of $\cal V$.\\
        ${\cal T}^{'} \gets$ front $(1-r_d)\times\left|{\cal T}^{'}\right|$ samples.
        }
        \If{$e = E_s$}{
        ${\cal T}^{'} \gets \cal T$.
        }
    }
    \KwOutput{${\cal T}^{'}$.}
\end{algorithm}

\textbf{For (2)}, we observe that the wide range of LiDAR (100m) and its high frequency (10Hz) result in redundant data. To accelerate training, we propose an RSD technique, as shown in Alg.~\ref{alg:1}.

This is a hierarchical sample downsampling technique~\cite{qin2024infobatch} that divides the training process into four stages based on predefined epochs and a downsampling ratio. We define $E_1$ epochs for the initial stage, $E_2$ epochs for the second stage, and a final epoch $E_s$, with a downsampling ratio $r_d$. Below, we detail each stage.

\textbf{(a)} In the first stage, we optimize SCR using the full training set $\cal T$ over $E_1$ epochs. For each sample, we compute the median $L_1$ loss, denoted ${\cal L}_m$, choosing the median for its robustness to outliers.

\textbf{(b)} In the second stage, we assess sample convergence by calculating the variance $\cal V$ of ${\cal L}_m$ within a sliding window $S$. At epoch $E_1 + S$, we sort $\cal T$ by $\cal V$ in descending order, retaining the top $(1-r_d)\times \left|{\cal T} \right|$ samples as the downsampled set ${\cal T}^{'}$. High-variance samples are prioritized because they indicate slower convergence, requiring more training focus.

\textbf{(c)} In the third stage, we repeat the process on ${\cal T}^{'}$, computing the median loss and its variance within $W$. We then downsample further, selecting the top $(1-r_d)\times \left|{\cal T} \right|$ samples, reducing the set to $\left(1-r_d\right)^2 \times \left|{\cal T} \right|$.

\textbf{(d)} In the final stage, we train on the full set $\cal T$ from epoch $E_s$ onward to ensure convergence of all samples.

With RSD, by simply computing loss variance within a window with minimal overhead, allows the network to exclude well-converged samples during training. It accelerates training while preserving accuracy, and experimental results demonstrate its effectiveness.

\noindent \textbf{Loss Function}.
In the scene-specific prediction heads training stage, LightLoc needs to train a sample classification network and an SCR. To optimize the classification network, we use cross-entropy loss with a smoothing factor of 0.1, as shown in Eq.~\ref{eq2}.
Then, we employ $L_1$ loss, as shown in Eq.~\ref{eq1}, to train an SCR.

\noindent \textbf{Inference}. In this stage, similar to previous work~\cite{Li_2023_CVPR, Yang_2024_CVPR}, LightLoc takes the query point cloud in the LiDAR coordinate as input, and regresses its correspondences in the world coordinate. Then, RANSAC is used to estimate the poses.

\subsection{Enhancing SLAM with SCG}\label{Sec.3.2}
The fast training in 5 minutes and confidence estimation capabilities of SCG enable its integration into SLAM~\cite{Zhang_2014_RSS} to help eliminate error accumulation. Below, we explore this application in detail.

We first modify the sample classification network into a hierarchical structure to obtain more detailed classification results. Specifically, we perform iterative clustering at the first level with $k_1$ clusters, and then apply a second-level clustering to obtain $k_2$ clusters. This results in $k_1*k_2$ clusters and their corresponding centers ${\{c_i\}_{i=1}^{k_1*k_2}}$. 
The confidence estimation is defined as the product of the probabilities from the two-level classification network.
Inspired by recent works~\cite{Dong_2021_CVPR, Li_2020_CVPR, Dong_2022_3DV}, we implement the classification networks as base and hyper networks. 
Notably, this hierarchical network can still be trained within 5 minutes.

In the process system, we use the position estimates provided by the SLAM~\cite{Zhang_2014_RSS}. Let ${\hat x}_t$ denote the estimation at time $t$, defined as follows:
\begin{equation}
\mathcal{N}(\hat x_{t}, \hat P_{t}) \sim \mathcal{N}\left(F_t x_{t-1} , F_t P_{t-1} F_{t}^{T} + W_{t}\right),
\label{eq3}
\end{equation}

\noindent where $F_t$ is the state transition matrix from $t-1$ to $t$ in SLAM.  ${\hat x}_t$ and $\hat{P}_t$ represent the prior state mean and covariance, respectively. 
$W_t$ is the process noise. 

The measurement system leverages the sample classification network.
Let $y_t$ denote the output cluster center from the classification network, the observed data from the measurement system are as follows:
\begin{equation}
z_{t} =y_t + v_t,
\label{eq4}
\end{equation}

\noindent where $v_t \sim \mathcal{N}(0, V_t)$ denotes the measurement noise with $V_t=I \times (1-c)$. Here, $I$ denotes the identity matrix, $c$ represents the confidence output from SCG, and $z_t$ is the noisy observed position distribution.

In the filter system, 
we use the measurement $z_t$ to update the state estimate through the Kalman filter:
\begin{equation}
\mathcal{N}(x_{t}, P_{t}) \sim \mathcal{N}\left(\hat x_t + K_t (z_t - \hat x_t), (I - K_{t}) \hat{P}_{t}\right),
\label{eq5}
\end{equation}

\noindent where $K_{t}=\frac{ \hat P_t}{V_t+ \hat P_t}$ is the Kalman gain. The posterior position estimate $x_t$ and covariance $P_{t}$ are updated, obtaining a corrected result.
\section{Experiment}
\subsection{Experimental Setup}
\begin{table*}[h!t]
\centering
\resizebox{1\linewidth}{!}{
\begin{tabular}{c|@{}l|cc|cccc|c}
\toprule

& \quad Methods & {{Training} {Time}} & {Params} & {15-13-06-37} & {17-13-26-39} &  {17-14-03-00} & {18-14-14-42} & {Average[m/$^{\circ}$]}\\


%
\midrule
& \quad PointLoc~\cite{Wang_2022_Sensors}  & \redcell 125h & \orangecell 13M & 10.75/2.36 & 11.07/2.21 & 11.53/1.92 & 9.82/2.07 & 10.79/2.14 \\
& \quad PosePN~\cite{Yu_2022_PR}  & \greencell 5h & \greencell 4M & 9.47/2.80 & 12.98/2.35 &  8.64/2.19 & 6.26/1.64 & 9.34/2.25 \\
& \quad PosePN++~\cite{Yu_2022_PR}  & \orangecell 11h & \greencell 5M & 4.54/1.83 & 6.44/1.78 & 4.89/1.55 & 4.64/1.61 & 5.13/1.69 \\
& \quad PoseMinkLoc~\cite{Yu_2022_PR}   & \greencell 7h & \orangecell 13M & 6.77/1.84 & 8.84/1.84 &  8.08/1.69 & 6.56/2.06 & 7.56/1.86 \\
& \quad PoseSOE~\cite{Yu_2022_PR} & \orangecell 15h & \greencell 5M & 4.17/1.76 & 6.16/1.81 &  5.42/1.87 & 4.16/1.70 & 4.98/1.79 \\

& \quad STCLoc~\cite{Yu_2022_TITS}   & \orangecell 20h & \greencell 9M & 5.14/1.27 & 6.12/1.21 & 5.32/\underline{1.08} & 4.76/1.19 & 5.34/1.19 \\
& \quad NIDALoc~\cite{Yu_2023_TITS} & \orangecell 20h & \greencell 8M & 3.71/1.50 & 5.40/1.40 & 3.94/1.30 & 4.08/1.30 & 4.28/1.38 \\
& \quad HypLiLoc~\cite{Wang_2023_CVPR}   & \orangecell 17h & \orangecell 52M & 5.03/1.46 & 4.31/1.43 & 3.61/1.11 & 2.61/\underline{1.09} & 3.89/1.27 \\
\multirow{-9}{*}{\rotatebox{90}{{APR}}} 
& \quad DiffLoc~\cite{Li_2024_CVPR} & \redcell 145h & \orangecell 40M & 2.03/\textbf{1.04} & 1.78/\textbf{0.79} & 2.05/\textbf{0.83} & 1.56/\textbf{0.83} & 1.86/\textbf{0.87} \\
\midrule

& \quad SGLoc~\cite{Li_2023_CVPR}   & \redcell 50h & \redcell 105M & 1.79/1.67 & 1.81/1.76 & 1.33/1.59 & 1.19/1.39 & 1.53/1.60 \\
\multirow{-1.5}{*}{\rotatebox{90}{{SCR}}} 
& \quad LiSA~\cite{Yang_2024_CVPR}  & \redcell 53h & \redcell 105M & \underline{0.94}/\underline{1.10} & \underline{1.17}/1.21 & \underline{0.84}/1.15 & \underline{0.85}/1.11 & \underline{0.95}/1.14 \\
& \quad LightLoc  & \greencell 1h &  \orangecell 22M & \textbf{0.82}/1.12 & \textbf{0.85}/\underline{1.07} & \textbf{0.81}/1.11 & \textbf{0.82}/1.16 & \textbf{0.83}/\underline{1.12} \\
\bottomrule
\end{tabular}
}
\caption{\textbf{Quantitative results on the QEOxford dataset.} We report the mean position [m] and orientation [$^{\circ}$]. Best results in \textbf{bold}, second best results \underline{underlined}. We list the training time and the parameters.}
\label{tab:1}
\end{table*}

\noindent \textbf{Dataset and Metrics.}
Following~\cite{Li_2023_CVPR, Yang_2024_CVPR}, we evaluate LightLoc for LiDAR localization on three large-scale outdoor benchmark datasets: Oxford Radar RobotCar~\cite{Dan_2020_ICRA}, QEOxford~\cite{Li_2023_CVPR} and NCLT~\cite{Carlone_2016_IJRR}. The mean position and orientation error are used as the evaluation metric.

\textbf{Oxford Radar RobotCar} (Oxford) is collected by sensors on an autonomous-capable Nissan LEAF platform, designed for urban scene localization tasks. 
Each trajectory spans approximately 10 km, covering an area of about 2 km$^2$. 
Point clouds are generated by dual Velodyne HDL-32E LiDAR sensors, with ground truth poses provided by a GPS/INS system. 
This dataset is captured under various weather and traffic conditions. Its diversity makes it highly suitable for comprehensive method evaluation. 


\textbf{QEOxford} is a quality-enhanced version of the Oxford dataset, with GPS/INS errors minimized through alignment techniques, which has been demonstrated to be beneficial for localization tasks~\cite{Li_2023_CVPR}.

\begin{table*}[h!t]
\centering
\resizebox{1\linewidth}{!}{
\begin{tabular}{c|@{}l|cc|cccc|c}
\toprule

& \quad Methods & {{Training} {Time}} & {Params} & {15-13-06-37} & {17-13-26-39} &  {17-14-03-00} & {18-14-14-42} & {Average[m/$^{\circ}$]}\\


%
\midrule
& \quad PointLoc~\cite{Wang_2022_Sensors}  & \redcell 125h & \orangecell 13M & 12.42/2.26 &  13.14/2.50 &  12.91/1.92 &  11.31/1.98 &  12.45/2.17 \\
& \quad PosePN~\cite{Yu_2022_PR}  & \greencell 5h & \greencell 4M & 14.32/3.06 &  16.97/2.49 &  13.48/2.60 &  9.14/1.78 &  13.48/2.48  \\
& \quad PosePN++~\cite{Yu_2022_PR} \  & \orangecell 11h & \greencell 5M & 9.59/1.92 &  10.66/1.92 &  9.01/1.51 &  8.44/1.71 &  9.43/1.77 \\
& \quad PoseMinkLoc~\cite{Yu_2022_PR}   & \greencell 7h & \orangecell 13M & 11.20/2.62 &  14.24/2.42 &  12.35/2.46 &  10.06/2.15 &  11.96/2.41 \\
& \quad PoseSOE~\cite{Yu_2022_PR} & \orangecell 15h & \greencell 5M & 7.59/1.94 &  10.39/2.08 &  9.21/2.12 &  7.27/1.87 &  8.62/2.00 \\

& \quad STCLoc~\cite{Yu_2022_TITS}   & \orangecell 20h & \greencell 9M & 6.93/1.48 &  7.55/\underline{1.23} &  7.44/1.24 &  6.13/1.15 &   7.01/1.28 \\
& \quad NIDALoc~\cite{Yu_2023_TITS} & \orangecell 20h & \greencell 8M & 5.45/1.40 &  7.63/1.56 &  6.68/1.26 &  4.80/1.18 &   6.14/1.35\\

& \quad HypLiLoc~\cite{Wang_2023_CVPR}   & \orangecell 17h & \orangecell 52M & 6.88/\underline{1.09} &  6.79/1.29 &  5.82/\underline{0.97} &  3.45/\underline{0.84} &   5.74/\underline{1.05}\\
\multirow{-9}{*}{\rotatebox{90}{{APR}}} 
& \quad DiffLoc~\cite{Li_2024_CVPR} & \redcell 145h & \orangecell 40M & 3.57/\textbf{0.88} &  3.65/\textbf{0.68} &  4.03/\textbf{0.70} &  2.86/\textbf{0.60} &   3.53/\textbf{0.72}\\
\midrule
& \quad SGLoc~\cite{Li_2023_CVPR}   & \redcell 50h & \redcell 105M & 3.01/1.91 &  4.07/2.07 &  3.37/1.89 &  2.12/1.66 &  3.14/1.88 \\
\multirow{-1.5}{*}{\rotatebox{90}{{SCR}}} 
& \quad LiSA~\cite{Yang_2024_CVPR}  & \redcell 53h & \redcell 105M & \underline{2.36}/1.29 & \underline{3.47}/1.43 & \underline{3.19}/1.34 & \textbf{1.95}/1.23 & \underline{2.74}/1.32 \\
& \quad LightLoc  & \greencell 1h & \orangecell 22M & \textbf{2.33}/1.21 & \textbf{3.19}/1.34 & \textbf{3.11}/1.24 & \underline{2.05}/1.20 & \textbf{2.67}/1.25 \\

\bottomrule
\end{tabular}
}
\caption{\textbf{Quantitative results on the Oxford dataset.} We report the mean position [m] and orientation [$^{\circ}$]. Best results in \textbf{bold}, second best results \underline{underlined}. We list the training time and the parameters.}
\label{tab:2}
\end{table*}

\textbf{NCLT} is a campus-area localization dataset collected using sensors mounted on a Segway robotic platform.  Each trajectory spans approximately 5.5km and covers an area of about 0.45km$^2$. The point cloud is gathered by the Velodyne HDL-32E LiDAR. Ground truth is obtained by SLAM. 
It covers a diverse range of scenarios, including both outdoor and indoor environments with varying structural complexities, making it challenging for localization methods.


\noindent \textbf{Implementation Details.}
LightLoc is implemented with Pytorch~\cite{Paszke_2019_NIPS} and MinkowskiEngine~\cite{Choy_2019_CVPR}. 
For experiments on the Oxford and QEOxford datasets, the following configurations are applied. 
For SCG, the cluster labels $k_1$ and $k_2$ are set to 25 and 100, respectively. 
For RSD, the downsampling ratio $r_d$ is fixed at 0.25, with start and stop ratios $r_{st}$ and $r_{sp}$ configured at 0.25 and 0.85, respectively. 
To accelerate classification training, a 150MB buffer stores global features on the GPU, performing 50 epochs over the training data with a batch size of 512.
During the SCR training stage, the model is trained 25 epochs with a batch size of 256.
Optimization employs the AdamW optimizer~\cite{Loshchilov_2020_ICLR} with a learning rate ranging from $5e^{-4}$ and $5e^{-3}$, following a one-cycle schedule~\cite{Smith_2019_super}.
All experiments are conducted on an NVIDIA RTX 4090 GPU.
Further details can be found in the supplementary.

\noindent \textbf{Baselines and Comparisons.}
To validate the performance of LightLoc, we conduct a comparative analysis against several state-of-the-art regression-based LiDAR localization methods. For APR, PointLoc~\cite{Wang_2022_Sensors},  PosePN~\cite{Yu_2022_PR}, PosePN++~\cite{Yu_2022_PR}, PoseMinkLoc~\cite{Yu_2022_PR}, PoseSOE~\cite{Yu_2022_PR} and HypLiLoc~\cite{Wang_2023_CVPR} are chosen as comparison methods, which take a single-frame point cloud as input. Additionally, we compare against multi-frame APR methods, such as STCLoc~\cite{Yu_2022_TITS}, NIDALoc~\cite{Yu_2023_TITS}, and DiffLoc~\cite{Li_2024_CVPR}, with DiffLoc representing the current state-of-the-art in LiDAR-based APR. 
For SCR, we compare LightLoc to SGLoc~\cite{Li_2023_CVPR} and LiSA~\cite{Yang_2024_CVPR}, with LiSA being the leading LiDAR localization method in regression-based methods. 

\begin{table*}[h!t]
\centering
\resizebox{1\linewidth}{!}{
\begin{tabular}{c|@{}l|cc|cccc|c}
\toprule

& \quad Methods & {{Training} {Time}} & {Params} & {2012-02-12} & {2012-02-19} &  {2012-03-31} & {2012-05-26$^\dag$} & {Average[m/$^{\circ}$]}\\


%
\midrule
& \quad PointLoc~\cite{Wang_2022_Sensors}  & \redcell 85h & \orangecell 13M & 7.23/4.88 &  6.31/3.89 &  6.71/4.32 &  9.55/5.21 &  7.45/4.58 \\
& \quad PosePN~\cite{Yu_2022_PR}  & \greencell 3h & \greencell 4M & 9.45/7.47 &  6.15/5.05 &  5.79/5.28 &  12.32/7.42 &   8.43/6.31 \\
& \quad PosePN++~\cite{Yu_2022_PR} \  & \greencell 8h &  \greencell 5M &4.97/3.75 &  3.68/2.65 &  4.35/3.38 &  8.42/4.30 &  5.36/3.52 \\
& \quad PoseMinkLoc~\cite{Yu_2022_PR}   & \greencell 5h & \orangecell 13M & 6.24/5.03 &  4.87/3.94 &  4.23/4.03 &  9.32/6.11 &   6.17/4.78 \\
& \quad PoseSOE~\cite{Yu_2022_PR} & \orangecell 10h & \greencell 5M & 13.09/8.05 &  6.16/4.51 &  5.24/4.56 &  13.27/7.85 &  9.44/6.24 \\

& \quad STCLoc~\cite{Yu_2022_TITS}   & \orangecell 14h & \greencell 9M & 4.91/4.34 &  3.25/3.10 &  3.75/4.04 &  7.53/4.95 &  4.86/4.11 \\
& \quad NIDALoc~\cite{Yu_2023_TITS} & \orangecell 14h & \greencell 8M & 4.48/3.59 &  3.14/2.52 &  3.67/3.46 &  6.32/4.67 &  4.40/3.56 \\

& \quad HypLiLoc~\cite{Wang_2023_CVPR}   & \orangecell 12h & \orangecell 52M & 1.71/3.56 &  1.68/2.69 &  1.52/2.90 &  \underline{2.29}/3.34 &  1.80/3.12  \\
\multirow{-9}{*}{\rotatebox{90}{{APR}}} 
& \quad DiffLoc~\cite{Li_2024_CVPR} & \redcell 100h & \orangecell 40M & \underline{0.99}/\underline{2.40} &  0.92/\underline{2.14} &  0.98/\underline{2.27} &  \textbf{1.36}/\textbf{2.48} &  \textbf{1.06}/\underline{2.32} \\
\midrule
& \quad SGLoc~\cite{Li_2023_CVPR}   & \redcell 42h & \redcell 105M & 1.20/3.08 &  1.20/3.05 &  1.12/3.28 &  3.48/4.43 &  1.75/3.46 \\
\multirow{-1.5}{*}{\rotatebox{90}{{SCR}}} 
& \quad LiSA~\cite{Yang_2024_CVPR}  & \redcell 44h & \redcell 105M & \textbf{0.97}/\textbf{2.23} & \underline{0.91}/\textbf{2.09} & \underline{0.87}/\textbf{2.21} & 3.11/\underline{2.72} & 1.47/\textbf{2.31}  \\
& \quad LightLoc  & \greencell 1h & \orangecell 23M & \underline{0.98}/2.76 & \textbf{0.89}/2.51 & \textbf{0.86}/2.67 & 3.10/3.26 & \underline{1.46}/2.80 \\

\bottomrule
\end{tabular}
}
\caption{\textbf{Quantitative results on the NCLT dataset.} We report the mean position [m] and orientation [$^{\circ}$]. Best results in \textbf{bold}, second best results \underline{underlined}. We list the training time and the parameters. $^\dag$ indicates that we discard areas with localization failure, as regression-based methods cannot generalize to unknown regions. More details can be found in the supplementary.}
\label{tab:3}
\end{table*}

\subsection{Comparison with State-of-the-art Methods}
\noindent \textbf{Results on Oxford.}  
We first evaluate the proposed LightLoc on the QEOxford, as shown in Tab.~\ref{tab:1}. 
We report the training time and parameters to assess computational efficiency, while mean position and orientation errors across test trajectories are used to evaluate localization accuracy.
LightLoc achieves an average error of 0.83m/1.12$^{\circ}$, which is the best and second best result in the position and orientation, respectively. Additionally, we are the only method that achieves training in 1 hour, with 22M parameters. Compared to the state-of-the-art method, LiSA, we get a comparable performance with about $50\times$ training time reduction and $4\times$ parameters reduction. While DiffLoc achieves higher orientation accuracy, 0.87$^{\circ}$ vs. 1.12$^{\circ}$, LightLoc offers the following advantages: (1) it requires only a single-frame input to output results, (2) position accuracy improves by 55.4$\%$, and (3) training time is $145\times$ faster.

Tab.~\ref{tab:2} presents the evaluation results of LightLoc on the (non-enhanced) Oxford dataset. Our method achieves an average error of 2.67m/1.25$^{\circ}$, ranking first in position and third in orientation. On this dataset, LightLoc maintains a 1-hour training time, achieving an excellent balance between accuracy and efficiency.

\noindent \textbf{Results on NCLT.}
We also evaluate our method on the NCLT dataset. Tab.~\ref{tab:3} summarizes the results of all methods with training time, parameters, and mean position/orientation error. 
LightLoc is the only method that achieves training within 1 hour. Additionally, LightLoc achieves an average error of 1.46m/2.80$^{\circ}$, ranking second and third in position and orientation, respectively.

It is important to note that our work is predominately concerned with the time needed for training, and, secondly, the storage demand of the parameters. 
While our method may not always achieve the highest performance on all test trajectories, it consistently delivers results within 1 hour of training on large-scale datasets, which is significantly faster than current state-of-the-art methods. 
LightLoc achieves a favorable balance between training time and performance, bridging the gap for the practical application of such methods in time-sensitive system upgrades, such as autonomous driving, drones, robotics, \emph{etc}.

\begin{figure}[t]
  \centering
  \includegraphics[width=1\columnwidth]{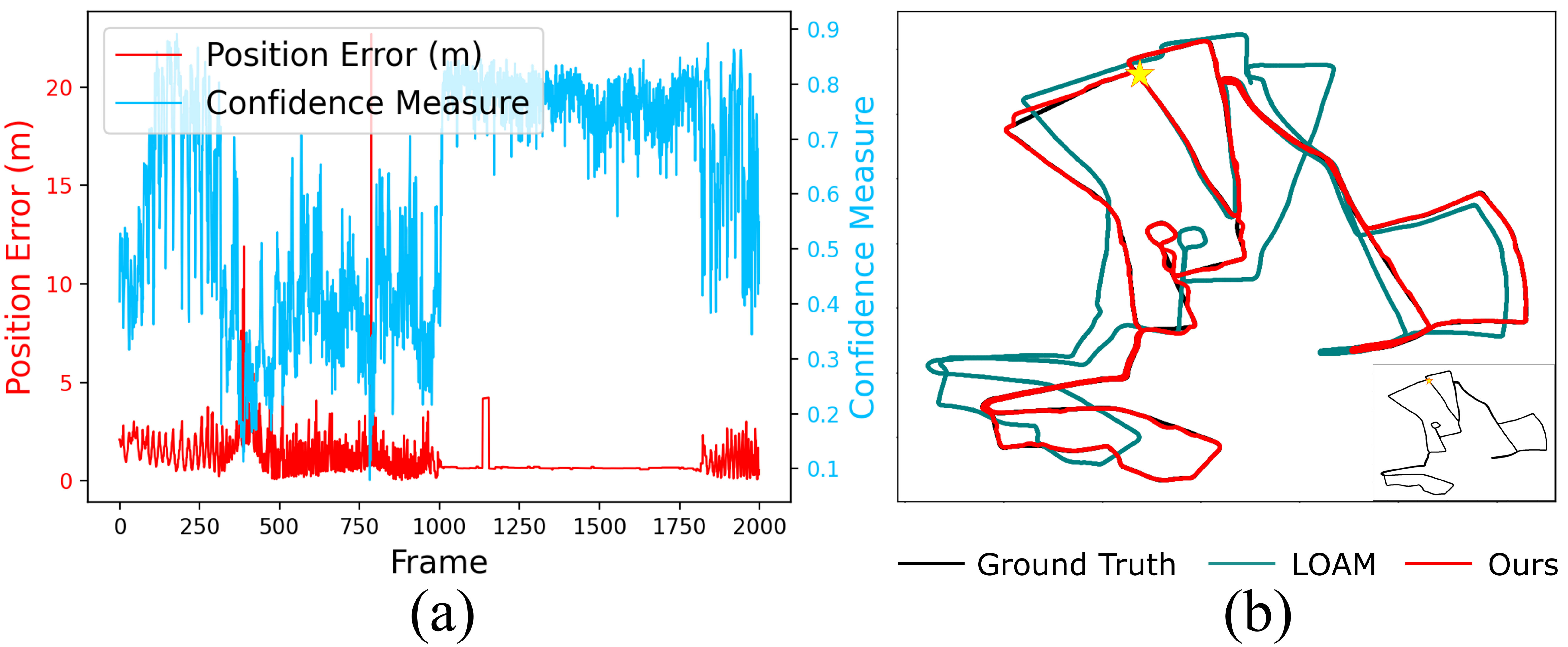}
   \vspace{-0.6cm}
   \caption{
   \textbf{Illustration of results with error accumulation eliminated}. (a) Relationship between position error and confidence estimation. (b) Visualization results of LOAM and ours. The star denotes the first frame. 
   }
   \label{fig:3}
\end{figure}
\noindent \textbf{Results with Error Accumulation Eliminated.}
We show that the sample classification guidance, which requires only 5 minutes to train, can be integrated into SLAM to eliminate error accumulation. As shown in Fig.~\ref{fig:3}, this is a plot of the relationship between the position error and the confidence estimation, demonstrating that they are highly correlated. 
Then, on the QEOxford dataset, we show the results of fusing them with SLAM, LOAM~\cite{Zhang_2014_RSS} in this paper, using Kalman filtering.
It is clear that the LOAM results deviate significantly from the true positions as the distance increases. In contrast, our method effectively corrects these errors, achieving highly accurate localization. We also present quantitative comparisons: the mean position error for LOAM is 87.81m, whereas our method reduces it to 4.48m, resulting in an improvement of 94.9$\%$.. This experiment clearly demonstrates that our scene classification guidance can be effectively incorporated into SLAM to eliminate error accumulation.

\noindent \textbf{Speed.}
Runtime time is a crucial metric in the LiDAR localization task. On the Oxford and NCLT datasets, the LiDAR scanning frequencies are 20Hz and 10Hz, respectively. Therefore, for real-time performance, the runtime must be less than 50ms and 100ms on these datasets.
The average runtime of LightLoc on the Oxford and NCLT datasets is 29ms and 48ms, respectively (with a batch size of 1), meeting the real-time speed requirements. Note that these runtimes include both regression and RANSAC solving, highlighting the method’s computational efficiency.

\begin{table}[t]
		\centering
		\resizebox{1\linewidth}{!}
  {
		\begin{tabular}{c|ccc|cc}
			\toprule
			             & Backbone            &  SCG         & RSD     & Training Time     & Mean Error~[m/$^{\circ}$]\\
			\midrule
			1            &               &               &               & 5.7h                 & 1.02/1.23\\
			2            & $\checkmark$  &               &               & 2.4h                 & 1.02/1.33\\
			3            & $\checkmark$    & $\checkmark$	  & 	 	      & 1.3h            & 0.82/1.10\\
			4            & $\checkmark$    &               & $\checkmark$    & 2h             & 1.03/1.36\\
            5            & $\checkmark$    & $\checkmark$    & $\checkmark$    & 1h            & 0.83/1.12\\
			\bottomrule
		\end{tabular}
	}
	\caption{\textbf{Ablation study}. We report the training time required to achieve comparable accuracy across different settings on the QEOxford dataset. \textbf{Backbone}: Scene-agnostic backbone used for feature learning. \textbf{SCG}: Sample classification guidance applied to help regression learning. \textbf{RSD}: Redundant sample downsampling is used to remove well-learned samples during training.}
    \label{tab:4}
\end{table}

\subsection{Ablation Study}
\noindent \textbf{Study on Backbone.}
As shown in Tab.~\ref{tab:4} between Row 1 and Row 2, freezing the scene-agnostic backbone and training only the scene-specific prediction heads results in significant time savings. It reduces training time by approximately 57.9$\%$ while maintaining similar performance. This study demonstrates that our scene-agnostic backbone, trained on 350K samples from the nuScenes dataset~\cite{Caesar_2020_CVPR}, generalizes well to new scenes. 
This discovery is encouraging, as it suggests that we can collect more data to train the scene-agnostic backbone to improve its generalizability.

\noindent \textbf{Study on SCG.}
We further conduct ablation experiments to demonstrate the importance of SCG. As shown in Tab.~\ref{tab:4}, the comparison between Row 2 and Row 3 shows that SCG improves position and orientation accuracy by 19.6$\%$ and 17.3$\%$, respectively. It also enhances training efficiency, reducing training time by approximately 45.8$\%$. 
This result demonstrates that the proposed SCG, which requires only 5 minutes of training for a new scene, can significantly enhance performance and reduce training time, effectively addressing the challenges of large coverage areas in large-scale outdoor scenes.

We examine the impact of different cluster numbers $k_1$ on the final performance, as shown in Tab.~\ref{tab:5}. 
Leveraging high accuracy and classification features (rather than classification results), our approach ensures robustness.

\noindent \textbf{Study on RSD.}
The results for RSD are shown in Row 4 and Row 5 of Tab.~\ref{tab:4}. Comparing Row 2 and Row 4, RSD reduces training time from 2.4 hours to 2 hours with minimal accuracy loss (1.02m/1.33$^\circ$ vs. 1.03m/1.36$^\circ$). 
Additionally, comparing Row 3 and Row 5, when using our method with guidance, the improvement is significant, boosting training efficiency by 23.1$\%$. 
With RSD, training time is further reduced, ultimately achieving the goal of completing training within 1 hour. This study demonstrates that RSD can significantly reduce training time without sacrificing accuracy, effectively addressing the challenge of handling large volumes of data in large-scale outdoor scenes. 

We also investigate the impact of different sampling strategies (\emph{e.g.}, random sampling, and uniform sampling) and sampling ratios on performance, as shown in Tab.~\ref{tab:6}. 
While uniform sampling yield comparable results at a $15\%$ ratio, further increasing the sampling ratio negatively impacts performance.
In contrast, our RSD shows no detectable accuracy drop at a pruning rate of $25\%$, and the drop remains minimal even at $35\%$. 
This is achieved by effectively selecting well-learned samples using the variance of the median loss, which ensures stable performance. 

\begin{table}[t]
	\resizebox{\linewidth}{!}{
	\begin{tabular}{@{}l|ccccc}
		\toprule
		Classes & 5 & 15 & 25  & 50          & 100            \\ \hline
		Accuracy [$\%$] &100 &99.86 & 99.79 & 99.55  & 99.19 \\ \hline
        Error [m/$^\circ$] & 0.90/1.18 & 0.85/1.12 & 0.83/1.12    & 0.89/1.17 & 0.89/1.16      \\
		\bottomrule
	\end{tabular}
	}
    \caption{\textbf{Effect of cluster number $k_1$.} We report classification accuracy [$\%$] and mean error [m/$^{\circ}$] on the QEOxford dataset.}
    \label{tab:5}
\end{table}

\begin{table}[t]
    \centering
	\resizebox{0.7\linewidth}{!}{
	\begin{tabular}{@{}l|ccc}
		\toprule
		Sampling & Random          & Uniform        & RSD           \\ \hline
		15$\%$  & 1.02/1.30 & 0.82/1.13   & 0.82/1.11     \\
        25$\%$  & 1.12/1.37 & 0.92/1.14   & 0.83/1.12     \\
        35$\%$  & 1.25/1.55 & 1.25/1.52   & 0.93/1.15     \\
		\bottomrule
	\end{tabular}
	}
    \caption{\textbf{Impact of different sampling strategies and ratios}. We report the mean error [m/$^{\circ}$] on the QEOxford dataset.}
    \label{tab:6}
\end{table}
\section{Conclusion and Future Work}
In this paper, we propose LightLoc, a novel LiDAR localization method capable of learning a new large-scale outdoor scene within 1 hour, while maintaining state-of-the-art performance.
Our method is based on freezing a scene-agnostic feature backbone and training only scene-specific prediction heads, enabling rapid learning of new scenes.
Beyond this design, We propose SCG and RSD, to address the challenges of large coverage areas and vast amounts of data in large-scale outdoor scenes, respectively. SCG uses a classification task to produce a sample probability distribution in 5 minutes, boosting regression training and accuracy, while RSD employs median loss variance to selectively reduce samples, enhancing training efficiency without sacrificing precision.
In addition, we show that SCG can be incorporated into SLAM to correct errors. 
Extensive experiments demonstrate the effectiveness of our methods.

In future work, we will explore incorporating data across more scenes, platforms, and LiDAR types to jointly train the backbone, further improving its generalizability.

\PAR{Acknowledgements} This work was partially supported by the Natural Science Foundation of China (No. 62171393),  the Fundamental Research Funds for the Central Universities (No.20720220064, No. 20720230033),  PDL (2022-PDL-12) and Xiaomi Young Talents Program.

\newpage
\appendix
\section*{Supplementary Material}

In this supplementary, we first describe the architecture and training process of the scene-agnostic feature backbone~(Sec.~\ref{sup_sec1}). Then, we describe the architecture and training process of the scene-specific head~(Sec.~\ref{sup_sec2}). We further provide additional results~(Sec.~\ref{sup_sec3}). Finally, we show more visualizations on the QEOxford~\cite{Dan_2020_ICRA, Li_2023_CVPR}, Oxford~\cite{Dan_2020_ICRA}, and NCLT~\cite{Carlone_2016_IJRR} datasets~(Sec.~\ref{sup_sec4}).

\section{Scene-agnostic Feature Backbone}\label{sup_sec1}
\subsection{Backbone Architecture}
\begin{figure}[t]
  \centering
  \includegraphics[width=1\columnwidth]{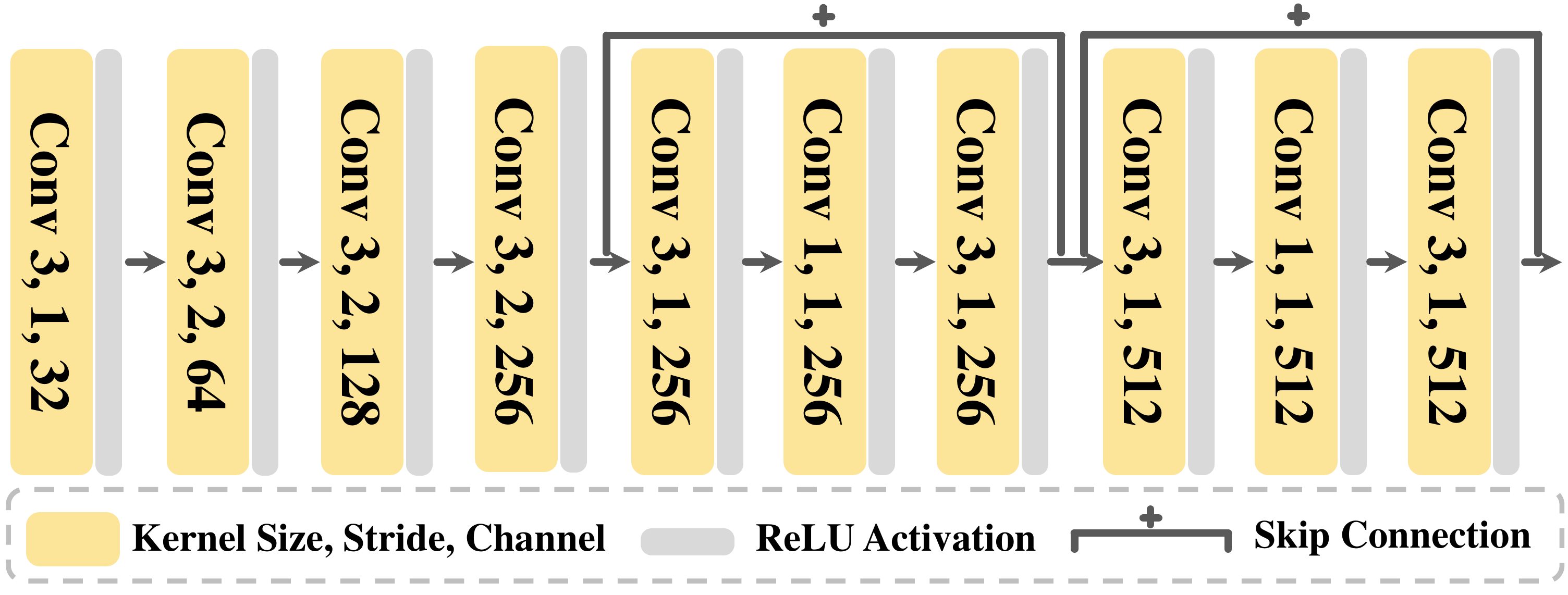}
   \caption{
   \textbf{Architecture of the scene-agnostic feature backbone}. The parameter count is about 16M.
   }
   \label{sup_fig1}
\end{figure}

Inspired by the DSAC$^{*}$~\cite{Brachmann_2021_PAMI}, we adopt the backbone of SGLoc~\cite{Li_2023_CVPR}, modifying it by reducing the feature dimension and the number of residual layers. As a result, the parameter count decreases significantly from 55M to 16M. The architecture of our backbone is illustrated in Fig.~\ref{sup_fig1}. 
The backbone processes a point cloud as input, progressively reducing the spatial resolution to $\frac{1}{8}$ while increasing the channel dimension to 512.

\subsection{Backbone Training}
We use the nuScenes dataset~\cite{Caesar_2020_CVPR}, including both Trainval and Test splits, totaling 350K samples, to train the scene-agnostic feature backbone.
The nuScenes dataset is collected using sensors mounted on two autonomous-capable Renault Zoe cars, equipped with identical sensor layouts, operating in Boston and Singapore—two cities known for their dense traffic and challenging driving conditions.
The point cloud is captured by a Velodyne HDL-32E LiDAR with a frequency of 20Hz.
The ground truth pose is obtained using an Advanced Navigation Spatial GPS$\&$IMU system, offering a position accuracy of 20mm, a heading accuracy of 0.2$^{\circ}$ with GNSS, and roll$\&$pitch accuracy of 0.1$^{\circ}$.

\begin{figure}[t]
  \centering
  \includegraphics[width=1\columnwidth]{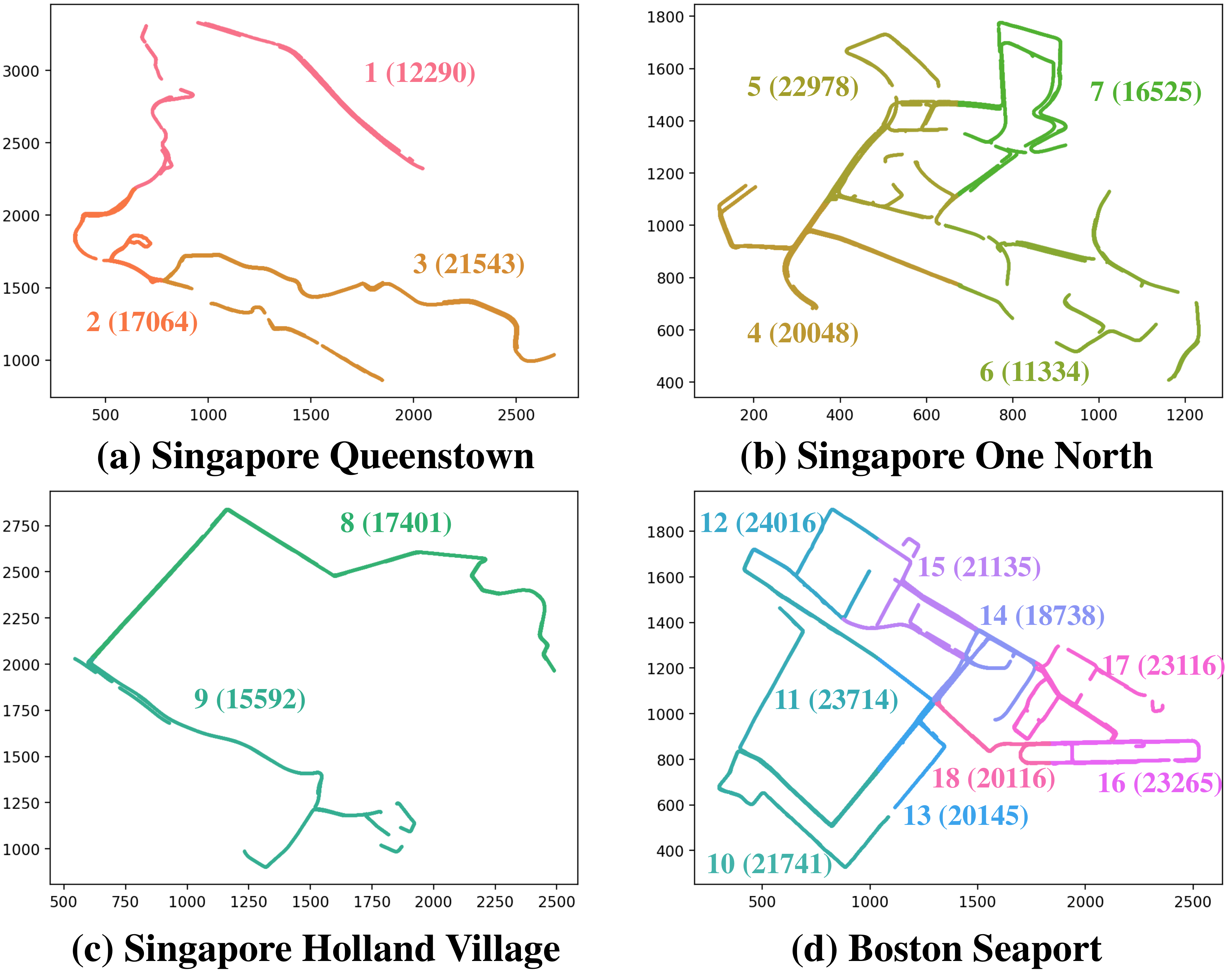}
   \caption{
   \textbf{Illustration of the multi-scene division in nuScenes.} The scene indices and the number of samples are reported.
   }
   \label{sup_fig2}
\end{figure}

We first apply the K-Means clustering and manually modify the results to divide the dataset into multi-scenes. The results are shown in Fig.~\ref{sup_fig2}, where we report the scene indices and the number of samples in each scene.
We train our backbone on 18 scenes in parallel, attaching 18 regression heads to it. Each regression head is a multi-layer perception with 6 layers and a width of 512. There is a skip connection after the first 3 layers of each head. We train the backbone with half-precision floating point weights. 
We perform random translation and rotation data augmentation on the input point clouds. Specifically, there is a 50$\%$ chance of translating along the x and y axes by -1 to 1 meters, rotating around the roll and pitch axes by -5$^{\circ}$ to 5$^{\circ}$, and rotating around the yaw axis by -10$^{\circ}$ to 10$^{\circ}$. 
It is important to note that, regardless of how point clouds are transformed, the learned ground truth 
remain consistent.

We train the backbone using a batch size of 32 samples per regression head. To prevent memory overflow, we process the forward and backward passes for 3 regression heads at a time. Gradients are accumulated across all 18 regression heads before performing a single parameter update.
\section{Scene-specific Head}\label{sup_sec2}
\subsection{Head Architecture}
In a new scene, we need to train a sample classification network and a SCR. Therefore, we have a classification head and a regression head. We now introduce each of these two heads in detail.

\begin{figure}[t]
  \centering
  \includegraphics[width=1\columnwidth]{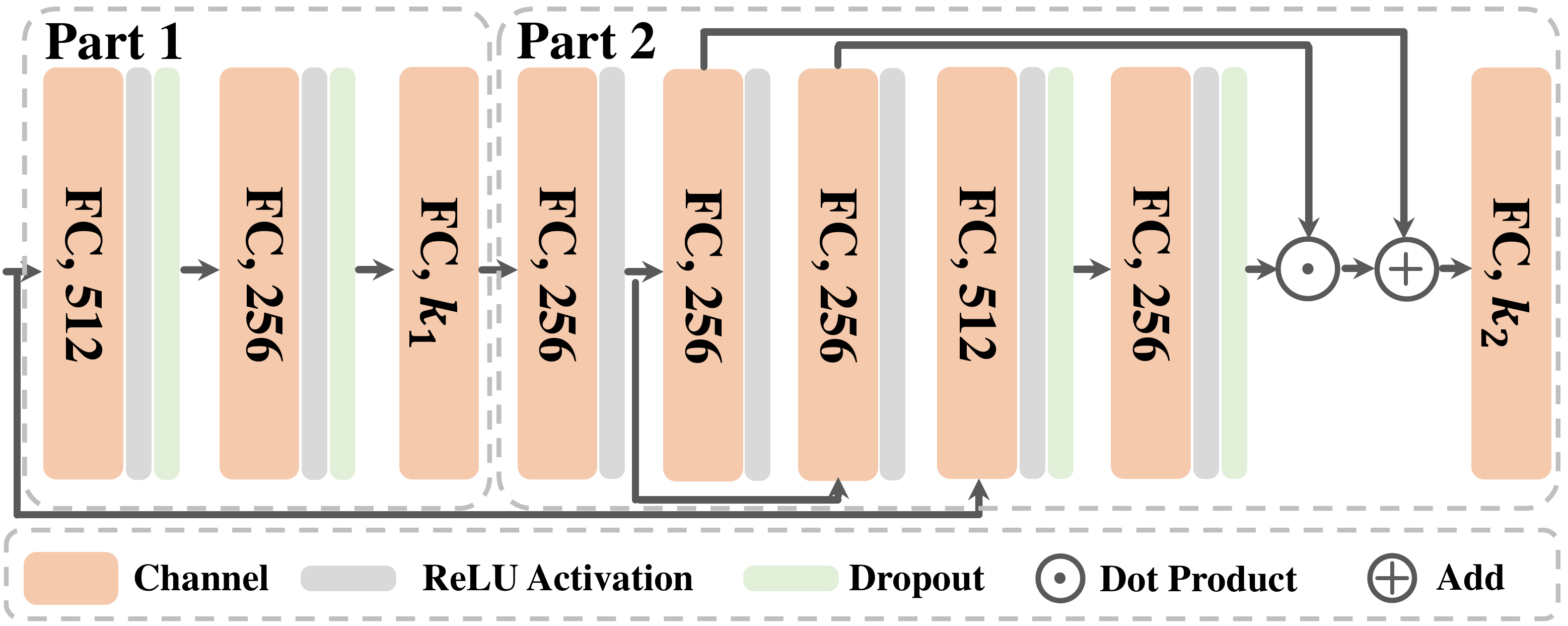}
   \caption{
   \textbf{Architecture of the sample classification head.} Part 1 is used for guiding SCR while Part 2 is used in extended applications requiring finer classification. This figure includes both parts to demonstrate the adaptability to different use cases. 
   }
   \label{sup_fig3}
\end{figure}

As we described in the main paper, inspired by recent work~\cite{Dong_2021_CVPR, Li_2020_CVPR, Dong_2022_3DV}, we implement the hierarchical classification network as the base and hyper network.
Fig.~\ref{sup_fig3} illustrates the complete sample classification head, designed to support both SCR and extended applications. 
The input is the feature after global max pooling.
As shown in Part 1, at the first level, the feature map from global max pooling is fed to an MLP to output the classification probability features $F_{k_1}$ with $k_1$ categories. 
Starting from the second level, as shown in Part 2, the feature pattern is modulated by a hyper network, according to the classification probability from the previous level. The intuition of the modulation~\cite{Perez_2018_AAAI} is that
similar feature patterns appearing in different regions should be classified under different labels. Then, an MLP is used to output the second level classification probability feature $F_{k_2}$ with $k_2$ categories.

In the sample classification guidance, only Part 1 is used to help SCR learning. Part 2 is employed only when addressing accumulated errors in SLAM, where finer classification results are required.

\begin{figure}[t]
  \centering
  \includegraphics[width=0.9\columnwidth]{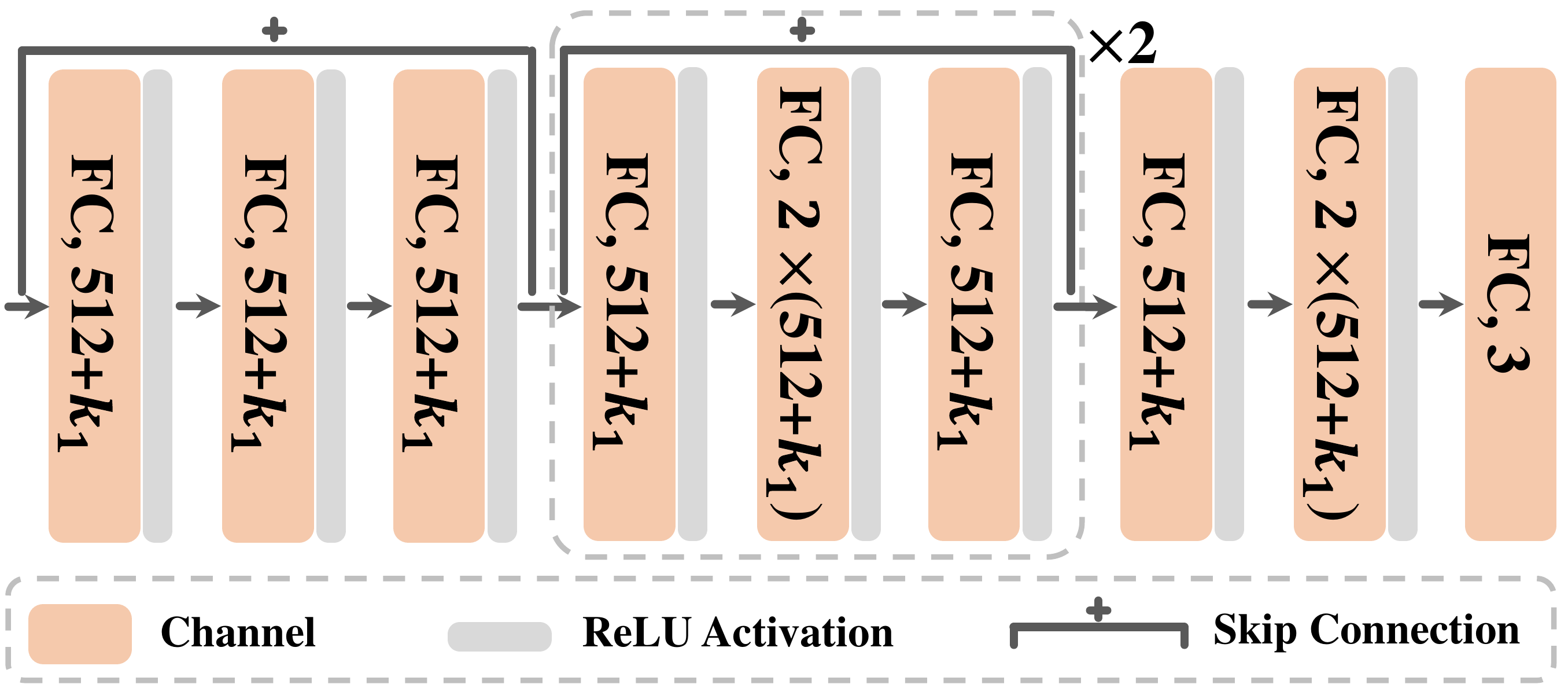}
   \caption{
   \textbf{Architecture of the regression head.} $k_1$ is the number of clusters. 
   }
   \label{sup_fig4}
\end{figure}

Fig.~\ref{sup_fig4} is the regression head, corresponding to the MLP in the SCR of Fig.~2 in the main paper. 
The input is the concatenated features: scene distribution features perturbed by Gaussian noise (standard deviation of 0.1) and dense descriptors obtained from the scene-agnostic backbone.
The features are transformed by a residual block, which is followed by $2$ sequential residual blocks. Finally, three FC layers are applied to get the corresponding point cloud in the world coordinates.

\subsection{Head Training}
In this paper, we train the scene-specific heads on QEOxford~\cite{Dan_2020_ICRA,Li_2023_CVPR}, Oxford~\cite{Dan_2020_ICRA} and NCLT~\cite{Carlone_2016_IJRR} datasets, respectively.

The \textbf{Oxford} dataset is collected in January 2019 along a central Oxford route, capturing a variety of weather (sunny, overcast) and lighting conditions (dim, glare) that make localization challenging. Following~\cite{Li_2023_CVPR, Yang_2024_CVPR}, we use data from 11-14-02-26, 14-12-05-52, 14-14-48-55, and 18-15-20-12 for training, and data from 15-13-06-37, 17-13-26-39, 17-14-03-00, and 18-14-14-42 for testing. The same trajectories are also selected from the \textbf{QEOxford} dataset, where GPS$\&$IMU errors are corrected using the PQEE~\cite{Li_2023_CVPR}.

The \textbf{NCLT} dataset is collected approximately biweekly from January 8, 2012 to April 5, 2013, on the University of Michigan’s North Campus. It includes a variety of environmental changes, such as seasonal variations, lighting conditions, and changes in building structures. The dataset also covers both indoor and outdoor scenes, which adds to the complexity. For our experiments, we use the data from 2012-01-22, 2012-02-02, 2012-02-18, and 2012-05-11 as the training set, and the data from 2012-02-12, 2012-02-19, 2012-03-31, and 2012-05-26 as the test set.

For classification head training, as mentioned in the main paper, we accelerate the process by creating a buffer on the GPU to store global features and their associated classification labels. The buffer is filled by cycling repeatedly through the shuffled training sequence. Each point cloud is augmented using a similar approach to the backbone training. The classification head is then trained by iterating over the shuffled buffer. Specifically, within a 5-minute training period (including the time spent filling the buffer), we complete 50 epochs.

For regression head training, we first apply the same data augmentation used in the backbone training to each frame of the input point cloud. Then, the point cloud is passed through the feature backbone to obtain dense descriptors. The features from the classification head are normalized to the unit sphere, Gaussian noise is added, and the features are normalized back to the unit sphere. Finally, the two feature sets are concatenated, and 256 voxels are randomly selected to learn their corresponding global coordinates through regression.
Throughout this process, we also apply the proposed redundant sample downsampling technique to enhance efficiency.

\subsection{Implementation Details}
For the Oxford and QEOxford datasets, we set the voxel size to 0.25. During classifier training, the number of clusters $k_1$ and $k_2$ are configured to 25 and 100, respectively. A 150MB buffer is constructed on the GPU to store features, enabling rapid training over 50 epochs within 5 minutes, following the ACE~\cite{Brachmann_2023_CVPR}. In the regressor training stage, the downsampling ratio $r_d$ in RSD, along with the start epoch $r_{st}$, stop epoch $r_{sp}$, and total training epochs $E$, are set to 0.25, 0.25, 0.85, and 25, respectively.

For the NCLT dataset, the voxel size, buffer size, and $k_1$ are set to 0.3, 120MB, and 100, respectively. In the regressor training phase, the $r_d$ and $E$ are configured to 0.15 and 30, respectively.
\section{Additional Results}\label{sup_sec3}
\subsection{Results of 2012-05-26 on NCLT}
As described in Tab.~3 of the main paper, we discard areas with localization failure, as regression-based methods cannot generalize to unknown regions. Details follow.

\begin{figure}[t]
  \centering
  \includegraphics[width=1\columnwidth]{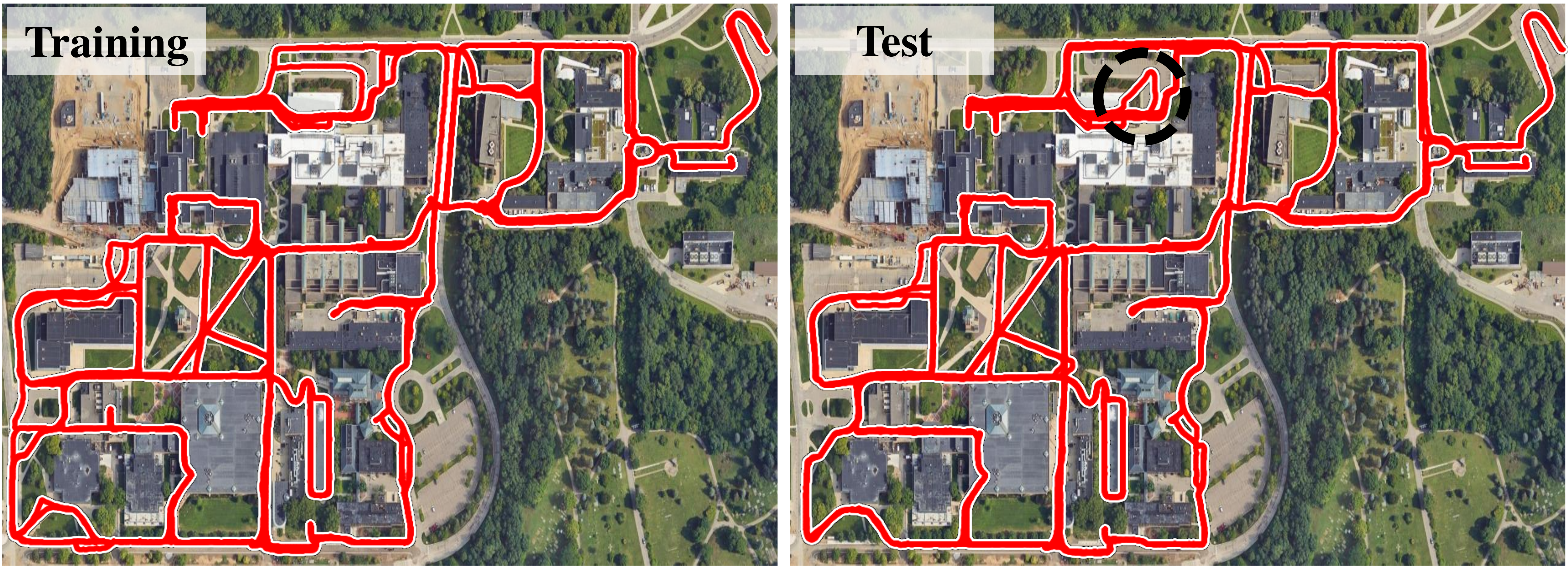}
   \caption{
   \textbf{Illustration of training and test trajectories of NCLT.} We use a black dashed circle to highlight the unknown region in the test trajectory.
   }
   \label{sup_fig5}
\end{figure}

\begin{table}[t]
	\resizebox{\linewidth}{!}
    {
	\begin{tabular}{@{}l|ccccc}
		\toprule
		Methods & HypLiLoc & DiffLoc & SGLoc  & LiSA          & Ours            \\ \hline
		Area Included  &2.90/3.47 & 1.88/2.43 & 3.81/4.74    & 3.30/2.84 & 3.81/3.48      \\
	    Area Excluded	 & 2.29/3.34 & 1.36/2.48 & 3.48/4.43    & 3.11/2.72 & 3.10/3.26 \\
        \bottomrule
	\end{tabular}
	}
    \caption{\textbf{Results on the 2012-05-26}. We report the mean error [m/$^{\circ}$] with unknown area included and excluded.}
    \label{sup_tab:1}
\end{table}

As shown in Fig.~\ref{sup_fig5}, we present the training and test trajectories of the NCLT dataset, highlighting the unknown region in 2012-05-26. Previous work~\cite{Sattler_2019_CVPR} demonstrates that regression-based methods are not guaranteed to generalize from the training data in practical scenarios. We also report the results with and without the area, as shown in Tab.~\ref{sup_tab:1}. It is clear that when the area is excluded, the errors of the different methods are significantly reduced. 
For a fair comparison of the methods, we report the results after excluding these regions, where the methods fail to provide useful information, and the results are essentially unreliable.

\subsection{Results of Training Backbone on KITTI}
In this section, we train the scene-agnostic feature backbone using the KITTI dataset~\cite{geiger2012_CVPR, Yu_2022_PR}. Since KITTI provides ground truth poses only for the training set (trajectories 00-10, totaling 23K samples), we use these 11 trajectories to train the backbone.
The KITTI dataset, collected in Karlsruhe, Germany, utilizes the autonomous driving platform Annieway. It captures diverse real-world driving scenarios, including urban, rural, and highway environments. Point clouds are recorded using a Velodyne HDL-64E LiDAR operating at 10Hz, while ground truth poses are derived from a GPS$\&$IMU system.

Similar to the backbone training with the nuScenes dataset, as described in Sec.~\ref{sup_sec1}, we train the backbone across 11 scenes in parallel, attaching 11 regression heads to it.

Then, we follow Sec.~\ref{sup_sec2} to train the scene-specific prediction heads on the QEOxford, Oxford, and NCLT datasets.

Tab.~\ref{sup_tab:2}, Tab.~\ref{sup_tab:3}, and Tab.~\ref{sup_tab:4} present the comparison results of training the backbone on the nuScenes dataset, evaluated across three different datasets. The results clearly show a decreasing trend in performance. Specifically, on the QEOxford dataset, position and orientation accuracy decrease by 80.7$\%$ and 54.5$\%$, respectively. In the Oxford and NCLT datasets, the corresponding decreases are 38.2$\%$/110.4$\%$ and 77.7$\%$/53.2$\%$, respectively.
We conclude that this is primarily due to insufficient data and differences in LiDAR types. This motivates us to incorporate data from more scenes, platforms, and LiDAR types to jointly train the backbone in future work, further enhancing its generalization capabilities.

\begin{table}[t]
        \centering
	\resizebox{\linewidth}{!}
    {
	\begin{tabular}{@{}l|cccc|c}
		\toprule
		Methods & 15-13-06-37 & 17-13-26-39 & 17-14-03-00  & 18-14-14-42 & Average\\ \hline
		nuScenes  &0.82/1.12 & 0.85/1.07 & 0.81/1.11    & 0.82/1.16  & 0.83/1.12\\
	    KITTI	 & 1.56/1.79 & 1.60/1.73 & 1.16/1.69    & 1.69/1.70 & 1.50/1.73\\
        \bottomrule
	\end{tabular}
	}
    \caption{\textbf{Results on QEOxford dataset}. 
    We report the mean error [m/$^{\circ}$] for scene-agnostic feature backbone training on the nuScenes and KITTI datasets.}
    \label{sup_tab:2}
\end{table}

\begin{table}[t]
        \centering
	\resizebox{\linewidth}{!}
    {
	\begin{tabular}{@{}l|cccc|c}
		\toprule
		Methods & 15-13-06-37 & 17-13-26-39 & 17-14-03-00  & 18-14-14-42 & Average\\ \hline
		nuScenes  &2.33/1.21 & 3.19/1.34 & 3.11/1.24    & 2.05/1.20  & 2.67/1.25\\
	    KITTI	 & 3.49/2.71 & 4.15/2.79 & 3.70/2.59    & 3.40/2.43 & 3.69/2.63\\
        \bottomrule
	\end{tabular}
	}
    \caption{\textbf{Results on Oxford dataset}. 
    We report the mean error [m/$^{\circ}$] for scene-agnostic feature backbone training on the nuScenes and KITTI datasets.}
    \label{sup_tab:3}
\end{table}

\begin{table}[t]
        \centering
	\resizebox{\linewidth}{!}
    {
	\begin{tabular}{@{}l|cccc|c}
		\toprule
		Methods & 2012-02-12 & 2012-02-19 & 2012-03-31  & 2012-05-26$^\dag$ & Average\\ \hline
		nuScenes  &0.98/2.76 & 0.89/2.51 & 0.86/2.67    & 3.10/3.26  & 1.46/2.80\\
	    KITTI	 & 1.51/4.15 & 1.61/3.70 & 1.46/3.99    & 5.77/5.30 & 2.59/4.29\\
        \bottomrule
	\end{tabular}
	}
    \caption{\textbf{Results on NCLT dataset}. 
    We report the mean error [m/$^{\circ}$] for scene-agnostic feature backbone training on the nuScenes and KITTI datasets.. $^\dag$ indicates that we discard areas with localization failure, as regression-based methods cannot generalize to unknown regions.}
    \label{sup_tab:4}
\end{table}
\section{Visualization}\label{sup_sec4}
\begin{figure*}
  \centering
  \includegraphics[width=1\linewidth]{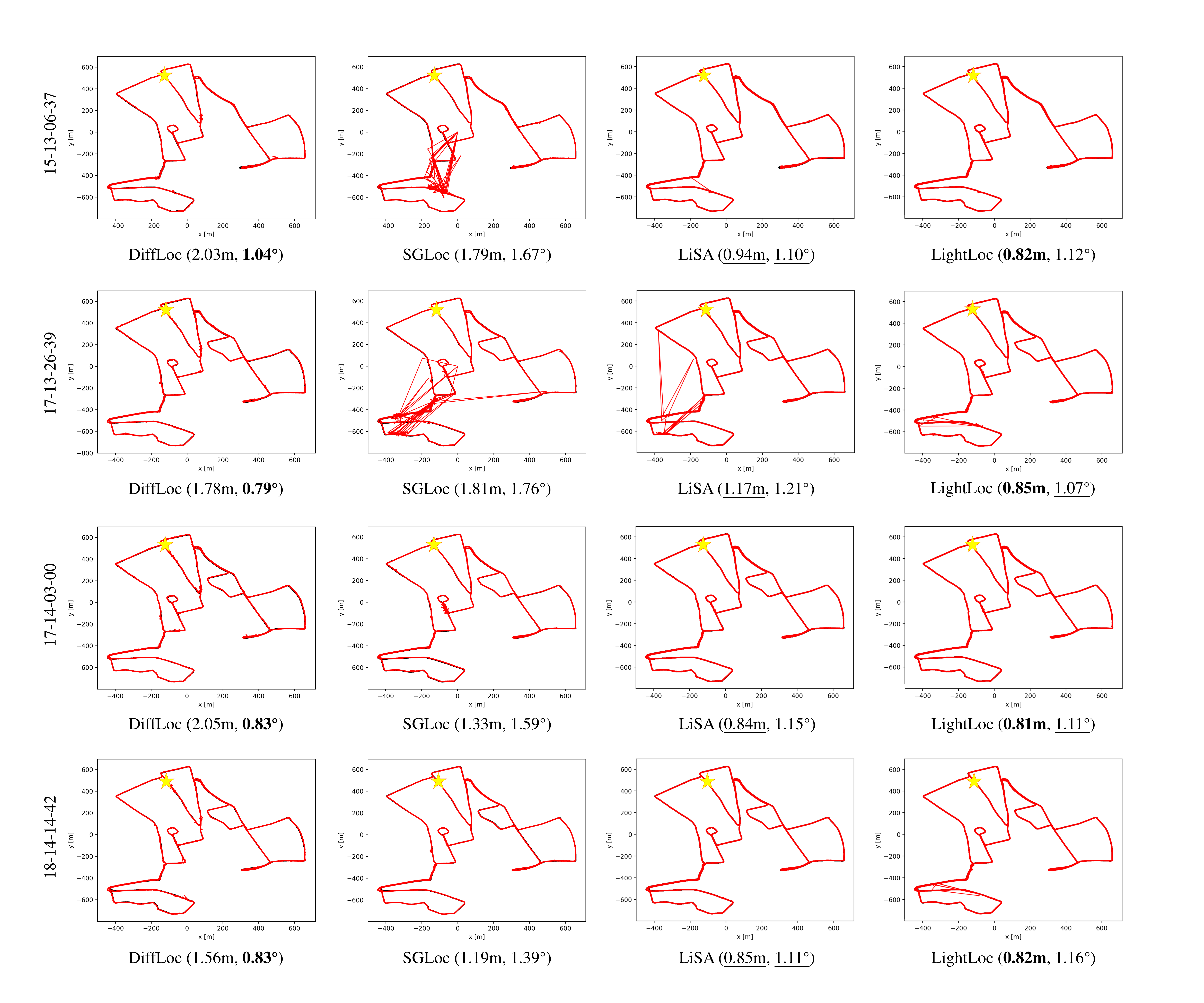}
  \caption{
LiDAR localization results on the QEOxford~\cite{Dan_2020_ICRA, Li_2023_CVPR} dataset. The ground truth and prediction are black and red lines, respectively. The
star denotes the first frame. The caption of each subfigure shows the mean position error (m) and orientation error ($^{\circ}$). For each trajectory, we highlight the \textbf{best} and \underline{second-best} results.
}
  \label{sup_fig6}
\end{figure*}

\begin{figure*}
  \centering
  \includegraphics[width=1\linewidth]{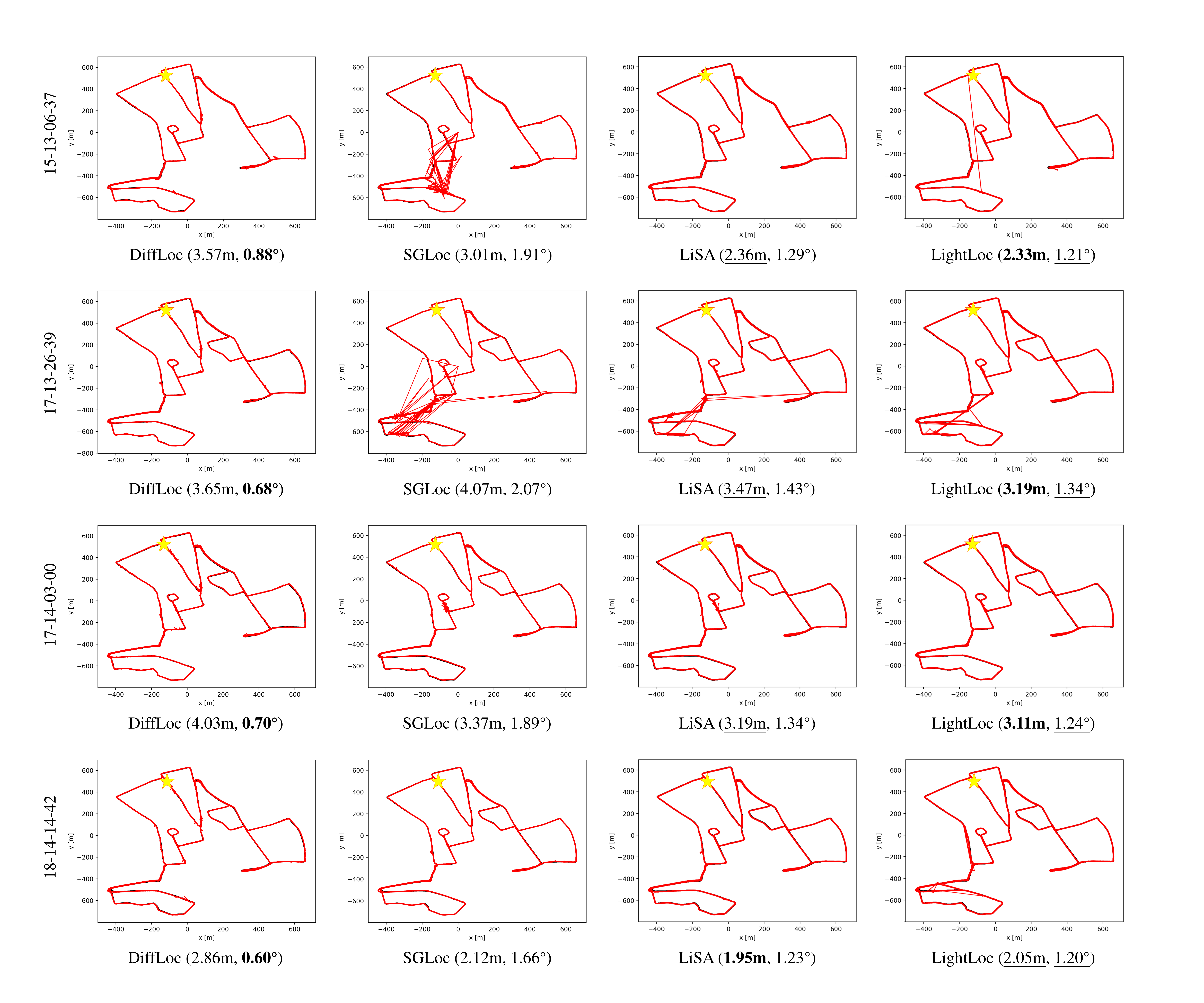}
  \caption{
LiDAR localization results on the Oxford~\cite{Dan_2020_ICRA} dataset. The ground truth and prediction are black and red lines, respectively. The
star denotes the first frame. The caption of each subfigure shows the mean position error (m) and orientation error ($^{\circ}$). For each trajectory, we highlight the \textbf{best} and \underline{second-best} results.
}
  \label{sup_fig7}
\end{figure*}

\begin{figure*}
  \centering
  \includegraphics[width=1\linewidth]{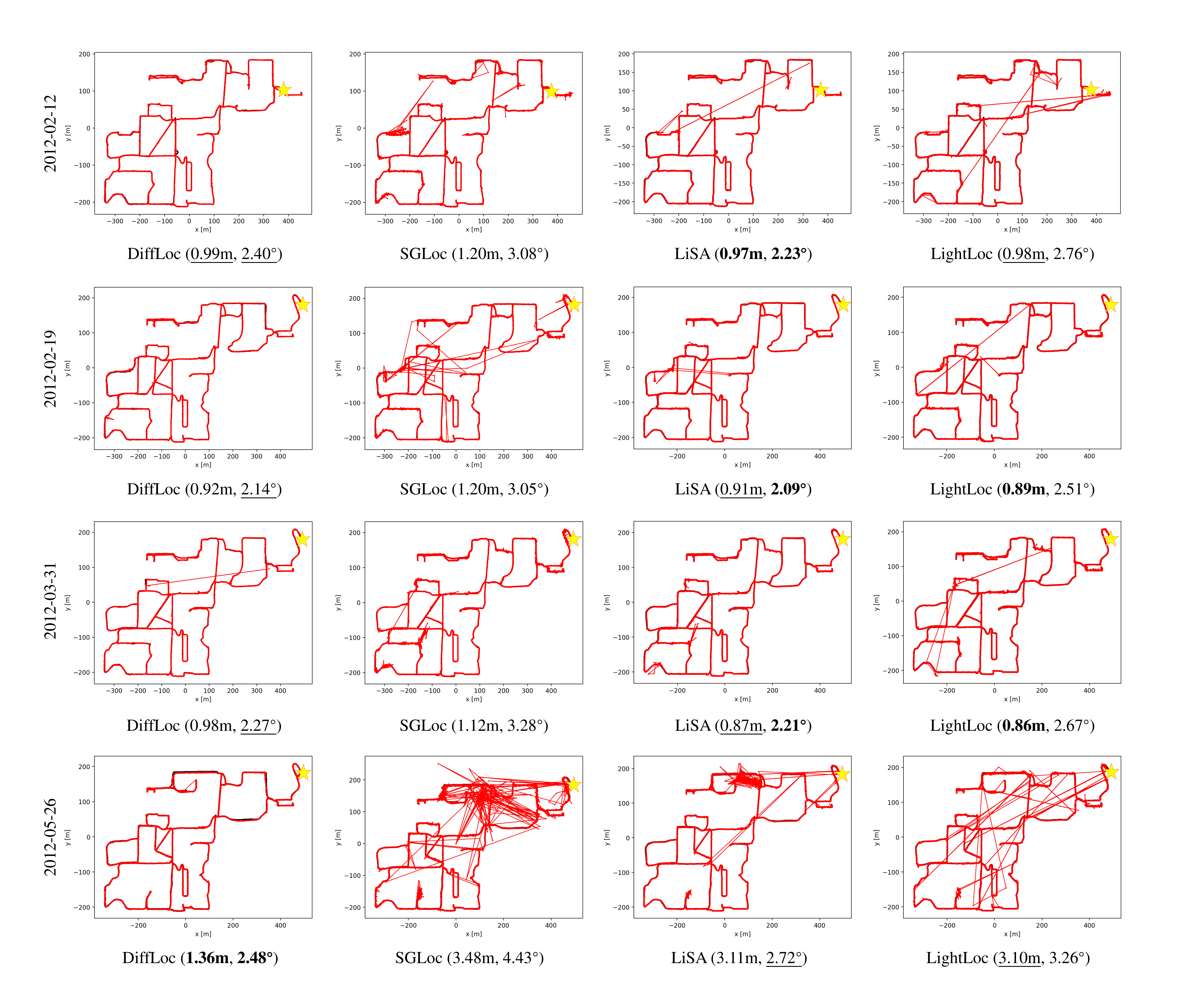}
  \caption{
LiDAR localization results on the NCLT~\cite{Carlone_2016_IJRR} dataset. The ground truth and prediction are black and red lines, respectively. The
star denotes the first frame. The caption of each subfigure shows the mean position error (m) and orientation error ($^{\circ}$). For each trajectory, we highlight the \textbf{best} and \underline{second-best} results.
}
  \label{sup_fig8}
\end{figure*}

We show more visualization results of the top 4 methods in the main paper (DiffLoc~\cite{Li_2024_CVPR}, SGLoc~\cite{Li_2023_CVPR}, LiSA~\cite{Yang_2024_CVPR}, and the proposed LightLoc) in Fig.~\ref{sup_fig6}, Fig.~\ref{sup_fig7}, and Fig.~\ref{sup_fig8} on the QEOxford, Oxford, and NCLT datasets, respectively.

Clearly, compared to the existing state-of-the-art method, LiSA, our predicted trajectories yield comparable results in terms of accuracy.

It is important to emphasize that our work primarily focuses on minimizing training time and reducing parameter storage requirements. While our method may not always outperform others on all test trajectories, it consistently achieves results within one hour of training on large-scale datasets, which is significantly faster than current state-of-the-art methods.
LightLoc achieves an effective balance between training time and performance, making it a practical solution for time-sensitive applications such as autonomous driving, drones, and robotics.

{
    \small
    \bibliographystyle{ieeenat_fullname}
    \bibliography{main}

\begin{thebibliography}{53}
\providecommand{\natexlab}[1]{#1}
\providecommand{\url}[1]{\texttt{#1}}
\expandafter\ifx\csname urlstyle\endcsname\relax
  \providecommand{\doi}[1]{doi: #1}\else
  \providecommand{\doi}{doi: \begingroup \urlstyle{rm}\Url}\fi

\bibitem[Ao et~al.(2023)Ao, Hu, Wang, Xu, and Guo]{Ao_2023_CVPR}
Sheng Ao, Qingyong Hu, Hanyun Wang, Kai Xu, and Yulan Guo.
\newblock Buffer: Balancing accuracy, efficiency, and generalizability in point cloud registration.
\newblock In \emph{CVPR}, pages 1255--1264, 2023.

\bibitem[Barnes et~al.(2020)Barnes, Gadd, Murcutt, Newman, and Posner]{Dan_2020_ICRA}
Dan Barnes, Matthew Gadd, Paul Murcutt, Paul Newman, and Ingmar Posner.
\newblock The oxford radar robotcar dataset: A radar extension to the oxford robotcar dataset.
\newblock In \emph{ICRA}, pages 6433--6438, 2020.

\bibitem[Brachmann and Rother(2021)]{Brachmann_2021_PAMI}
Eric Brachmann and Carsten Rother.
\newblock Visual camera re-localization from rgb and rgb-d images using dsac.
\newblock \emph{IEEE TPAMI}, 44\penalty0 (9):\penalty0 5847--5865, 2021.

\bibitem[Brachmann et~al.(2023)Brachmann, Cavallari, and Prisacariu]{Brachmann_2023_CVPR}
Eric Brachmann, Tommaso Cavallari, and Victor~Adrian Prisacariu.
\newblock Accelerated coordinate encoding: Learning to relocalize in minutes using rgb and poses.
\newblock In \emph{CVPR}, pages 5044--5053, 2023.

\bibitem[Brahmbhatt et~al.(2018)Brahmbhatt, Gu, Kim, Hays, and Kautz]{Brahmbhatt_2018_CVPR}
Samarth Brahmbhatt, Jinwei Gu, Kihwan Kim, James Hays, and Jan Kautz.
\newblock Geometry-aware learning of maps for camera localization.
\newblock In \emph{CVPR}, pages 2616--2625, 2018.

\bibitem[Caesar et~al.(2020)Caesar, Bankiti, Lang, Vora, Liong, Xu, Krishnan, Pan, Baldan, and Beijbom]{Caesar_2020_CVPR}
Holger Caesar, Varun Bankiti, Alex~H Lang, Sourabh Vora, Venice~Erin Liong, Qiang Xu, Anush Krishnan, Yu Pan, Giancarlo Baldan, and Oscar Beijbom.
\newblock nuscenes: A multimodal dataset for autonomous driving.
\newblock In \emph{CVPR}, pages 11621--11631, 2020.

\bibitem[Chen et~al.(2022)Chen, Li, Wang, and Prisacariu]{Chen_2022_ECCV}
Shuai Chen, Xinghui Li, Zirui Wang, and Victor Prisacariu.
\newblock Dfnet: Enhance absolute pose regression with direct feature matching.
\newblock In \emph{ECCV}, 2022.

\bibitem[Chen et~al.(2024{\natexlab{a}})Chen, Bhalgat, Li, Bian, Li, Wang, and Prisacariu]{Chen_2024_CVPRB}
Shuai Chen, Yash Bhalgat, Xinghui Li, Jia-Wang Bian, Kejie Li, Zirui Wang, and Victor~Adrian Prisacariu.
\newblock Neural refinement for absolute pose regression with feature synthesis.
\newblock In \emph{CVPR}, pages 20987--20996, 2024{\natexlab{a}}.

\bibitem[Chen et~al.(2024{\natexlab{b}})Chen, Cavallari, Prisacariu, and Brachmann]{Chen_2024_CVPR}
Shuai Chen, Tommaso Cavallari, Victor~Adrian Prisacariu, and Eric Brachmann.
\newblock Map-relative pose regression for visual re-localization.
\newblock In \emph{CVPR}, pages 20665--20674, 2024{\natexlab{b}}.

\bibitem[Choy et~al.(2019)Choy, Park, and Koltun]{Choy_2019_CVPR}
Christopher Choy, Jaesik Park, and Vladlen Koltun.
\newblock Fully convolutional geometric features.
\newblock In \emph{CVPR}, pages 8958--8966, 2019.

\bibitem[Dong et~al.(2021)Dong, Fan, Wang, Shi, Yi, Funkhouser, Chen, and Guibas]{Dong_2021_CVPR}
Siyan Dong, Qingnan Fan, He Wang, Ji Shi, Li Yi, Thomas Funkhouser, Baoquan Chen, and Leonidas~J Guibas.
\newblock Robust neural routing through space partitions for camera relocalization in dynamic indoor environments.
\newblock In \emph{CVPR}, pages 8544--8554, 2021.

\bibitem[Dong et~al.(2022)Dong, Wang, Zhuang, Kannala, Pollefeys, and Chen]{Dong_2022_3DV}
Siyan Dong, Shuzhe Wang, Yixin Zhuang, Juho Kannala, Marc Pollefeys, and Baoquan Chen.
\newblock Visual localization via few-shot scene region classification.
\newblock In \emph{3DV}, pages 393--402, 2022.

\bibitem[Fischler and Bolles(1981)]{Fischler_1981_Commun}
Martin~A. Fischler and Robert~C. Bolles.
\newblock Random sample consensus: A paradigm for model fitting with applications to image analysis and automated cartography.
\newblock \emph{Commun. ACM}, 15:\penalty0 381--395, 1981.

\bibitem[Geiger et~al.(2012)Geiger, Lenz, and Urtasun]{geiger2012_CVPR}
Andreas Geiger, Philip Lenz, and Raquel Urtasun.
\newblock Are we ready for autonomous driving? the kitti vision benchmark suite.
\newblock In \emph{CVPR}, pages 3354--3361, 2012.

\bibitem[Ilya and Hutter(2020)]{Loshchilov_2020_ICLR}
Loshchilov Ilya and Frank Hutter.
\newblock Decoupled weight decay regularization.
\newblock In \emph{ICLR}, 2020.

\bibitem[Jin et~al.(2024)Jin, Armeni, Pollefeys, and Barath]{Jin_2024_CVPR}
Shengze Jin, Iro Armeni, Marc Pollefeys, and Daniel Barath.
\newblock Multiway point cloud mosaicking with diffusion and global optimization.
\newblock In \emph{CVPR}, pages 20838--20849, 2024.

\bibitem[Kendall et~al.(2015)Kendall, Grimes, and Cipolla]{Kendall_2015_ICCV}
Alex Kendall, Matthew Grimes, and Roberto Cipolla.
\newblock Posenet: A convolutional network for real-time 6-dof camera relocalization.
\newblock In \emph{ICCV}, page 2938–2946, 2015.

\bibitem[Komorowski(2021)]{Komorowski_2021_WACV}
Jacek Komorowski.
\newblock Minkloc3d: Point cloud based large-scale place recognition.
\newblock In \emph{WACV}, pages 1790--1799, 2021.

\bibitem[Li et~al.(2023)Li, Yu, Wang, Hu, Shen, and Wen]{Li_2023_CVPR}
Wen Li, Shangshu Yu, Cheng Wang, Guosheng Hu, Siqi Shen, and Chenglu Wen.
\newblock Sgloc: Scene geometry encoding for outdoor lidar localization.
\newblock In \emph{CVPR}, pages 9286--9295, 2023.

\bibitem[Li et~al.(2024)Li, Yang, Yu, Hu, Wen, Cheng, and Wang]{Li_2024_CVPR}
Wen Li, Yuyang Yang, Shangshu Yu, Guosheng Hu, Chenglu Wen, Ming Cheng, and Cheng Wang.
\newblock Diffloc: Diffusion model for outdoor lidar localization.
\newblock In \emph{CVPR}, pages 15045--15054, 2024.

\bibitem[Li et~al.(2020)Li, Wang, Zhao, Verbeek, and Kannala]{Li_2020_CVPR}
Xiaotian Li, Shuzhe Wang, Yi Zhao, Jakob Verbeek, and Juho Kannala.
\newblock Hierarchical scene coordinate classification and regression for visual localization.
\newblock In \emph{CVPR}, pages 11983--11992, 2020.

\bibitem[Luo et~al.(2023)Luo, Zheng, Li, Fan, Yu, Cao, Li, and Shen]{Luo_2023_ICCV}
Lun Luo, Shuhang Zheng, Yixuan Li, Yongzhi Fan, Beinan Yu, Si-Yuan Cao, Junwei Li, and Hui-Liang Shen.
\newblock Bevplace: Learning lidar-based place recognition using bird's eye view images.
\newblock In \emph{ICCV}, pages 8700--8709, 2023.

\bibitem[M{\"u}ller et~al.(2019)M{\"u}ller, Kornblith, and Hinton]{Muller_2019_NIPS}
Rafael M{\"u}ller, Simon Kornblith, and Geoffrey~E Hinton.
\newblock When does label smoothing help?
\newblock \emph{NeurIPS}, 32, 2019.

\bibitem[Nguyen et~al.(2024)Nguyen, Fontan, Milford, and Fischer]{Nguyen_2024_WACV}
Son~Tung Nguyen, Alejandro Fontan, Michael Milford, and Tobias Fischer.
\newblock Focustune: Tuning visual localization through focus-guided sampling.
\newblock In \emph{WACV}, pages 3606--3615, 2024.

\bibitem[Nicholas et~al.(2015)Nicholas, Arash, and Ryan]{Carlone_2016_IJRR}
Carlevaris-Bianco Nicholas, K.~Ushani Arash, and M.~Eustice Ryan.
\newblock University of michigan north campus long-term vision and lidar dataset.
\newblock \emph{IJRR}, 35:\penalty0 545--565, 2015.

\bibitem[Paszke et~al.(2019)Paszke, Gross, Massa, Lerer, Bradbury, Chanan, Killeen, Lin, Gimelshein, Antiga, et~al.]{Paszke_2019_NIPS}
Adam Paszke, Sam Gross, Francisco Massa, Adam Lerer, James Bradbury, Gregory Chanan, Trevor Killeen, Zeming Lin, Natalia Gimelshein, Luca Antiga, et~al.
\newblock Pytorch: An imperative style, high-performance deep learning library.
\newblock \emph{NeurIPS}, 32, 2019.

\bibitem[Perez et~al.(2018)Perez, Strub, De~Vries, Dumoulin, and Courville]{Perez_2018_AAAI}
Ethan Perez, Florian Strub, Harm De~Vries, Vincent Dumoulin, and Aaron Courville.
\newblock Film: Visual reasoning with a general conditioning layer.
\newblock In \emph{AAAI}, 2018.

\bibitem[Qi et~al.(2017)Qi, Yi, Su, and Guibas]{Qi_2017_NIPS}
Charles~Ruizhongtai Qi, Li Yi, Hao Su, and Leonidas~J Guibas.
\newblock Pointnet++: Deep hierarchical feature learning on point sets in a metric space.
\newblock \emph{NeurIPS}, 30, 2017.

\bibitem[Qin et~al.(2022)Qin, Yu, Wang, Guo, Peng, and Xu]{Qin_2022_CVPR}
Zheng Qin, Hao Yu, Changjian Wang, Yulan Guo, Yuxing Peng, and Kai Xu.
\newblock Geometric transformer for fast and robust point cloud registration.
\newblock In \emph{CVPR}, pages 11143--11152, 2022.

\bibitem[Qin et~al.(2024)Qin, Wang, Zheng, Gu, Peng, Xu, Zhou, Shang, Sun, Xie, and You]{qin2024infobatch}
Ziheng Qin, Kai Wang, Zangwei Zheng, Jianyang Gu, Xiangyu Peng, Zhaopan Xu, Daquan Zhou, Lei Shang, Baigui Sun, Xuansong Xie, and Yang You.
\newblock Infobatch: Lossless training speed up by unbiased dynamic data pruning.
\newblock In \emph{ICLR}, 2024.

\bibitem[Sattler et~al.(2019)Sattler, Zhou, Pollefeys, and Leal-Taixe]{Sattler_2019_CVPR}
Torsten Sattler, Qunjie Zhou, Marc Pollefeys, and Laura Leal-Taixe.
\newblock Understanding the limitations of cnn-based absolute camera pose regressionr.
\newblock In \emph{CVPR}, pages 3302--3312, 2019.

\bibitem[Shavit et~al.(2021)Shavit, Ferens, and Keller]{Shavit_2021_ICCV}
Yoli Shavit, Ron Ferens, and Yosi Keller.
\newblock Learning multi-scene absolute pose regression with transformers.
\newblock In \emph{CVPR}, pages 2733--2742, 2021.

\bibitem[Smith and Topin(2019)]{Smith_2019_super}
Leslie~N Smith and Nicholay Topin.
\newblock Super-convergence: Very fast training of neural networks using large learning rates.
\newblock In \emph{AIMDO}, pages 369--386, 2019.

\bibitem[Uy and Lee(2018)]{Uy_2018_CVPR}
Mikaela~Angelina Uy and Gim~Hee Lee.
\newblock Pointnetvlad: Deep point cloud based retrieval for large-scale place recognition.
\newblock In \emph{CVPR}, pages 4470--4479, 2018.

\bibitem[Vaswani et~al.(2017)Vaswani, Shazeer, Parmar, Uszkoreit, Jones, Gomez, Kaiser, and Polosukhin]{Vaswani_2017_NIPS}
Ashish Vaswani, Noam Shazeer, Niki Parmar, Jakob Uszkoreit, Llion Jones, Aidan~N Gomez, {\L}ukasz Kaiser, and Illia Polosukhin.
\newblock Attention is all you need.
\newblock \emph{NeurIPS}, 30, 2017.

\bibitem[Wang et~al.(2020)Wang, Chen, Lu, Zhao, Trigoni, and Markham]{Wang_2020_AAAI}
Bing Wang, Chaohao Chen, Chrisxiaoxuan Lu, Peijun Zhao, Niki Trigoni, and Andrew Markham.
\newblock Atloc: Attention guided camera localization.
\newblock In \emph{AAAI}, pages 10393--10401, 2020.

\bibitem[Wang et~al.(2024{\natexlab{a}})Wang, Jiang, Galliani, Vogel, and Pollefeys]{Wang_2024_CVPR}
Fangjinhua Wang, Xudong Jiang, Silvano Galliani, Christoph Vogel, and Marc Pollefeys.
\newblock Glace: Global local accelerated coordinate encoding.
\newblock In \emph{CVPR}, pages 21562--21571, 2024{\natexlab{a}}.

\bibitem[Wang et~al.(2023{\natexlab{a}})Wang, Kang, She, Tay, Hartmannsgruber, and Navarro]{Wang_2023_AAAI}
Sijie Wang, Qiyu Kang, Rui She, Wee~Peng Tay, Andreas Hartmannsgruber, and Diego~Navarro Navarro.
\newblock Robustloc: Robust camera pose regression in challenging driving environments.
\newblock In \emph{AAAI}, pages 6209--6216, 2023{\natexlab{a}}.

\bibitem[Wang et~al.(2023{\natexlab{b}})Wang, Kang, She, Wang, Zhao, Song, and Tay]{Wang_2023_CVPR}
Sijie Wang, Qiyu Kang, Rui She, Wei Wang, Kai Zhao, Yang Song, and Wee~Peng Tay.
\newblock Hypliloc: Towards effective lidar pose regression with hyperbolic fusion.
\newblock In \emph{CVPR}, pages 5176--5185, 2023{\natexlab{b}}.

\bibitem[Wang et~al.(2024{\natexlab{b}})Wang, She, Kang, Jian, Zhao, Song, and Tay]{Wang_2024_AAAI}
Sijie Wang, Rui She, Qiyu Kang, Xingchao Jian, Kai Zhao, Yang Song, and Wee~Peng Tay.
\newblock Distilvpr: Cross-modal knowledge distillation for visual place recognition.
\newblock In \emph{AAAI}, pages 10377--10385, 2024{\natexlab{b}}.

\bibitem[Wang et~al.(2022)Wang, Wang, Zhao, Chen, Clark, Yang, Markham, and Trigoni]{Wang_2022_Sensors}
Wei Wang, Bing Wang, Peijun Zhao, Changhao Chen, Ronald Clark, Bo Yang, Andrew Markham, and Niki Trigoni.
\newblock Pointloc: Deep pose regressor for lidar point cloud localization.
\newblock \emph{IEEE Sensors}, 22:\penalty0 959--968, 2022.

\bibitem[Wang and Solomon(2019)]{Wang_2019_ICCV}
Yue Wang and Justin~M. Solomon.
\newblock Deep closest point: Learning representations for point cloud registration.
\newblock In \emph{ICCV}, pages 3523--3532, 2019.

\bibitem[Xia et~al.(2021)Xia, Xu, Li, Wang, Du, Cremers, and Stilla]{Xia_2021_CVPR}
Yan Xia, Yusheng Xu, Shuang Li, Rui Wang, Juan Du, Daniel Cremers, and Uwe Stilla.
\newblock Soe-net: A self-attention and orientation encoding network for point cloud based place recognition.
\newblock In \emph{CVPR}, pages 11348--11357, 2021.

\bibitem[Xia et~al.(2023)Xia, Gladkova, Wang, Li, Stilla, Henriques, and Cremers]{Xia_2023_ICCV}
Yan Xia, Mariia Gladkova, Rui Wang, Qianyun Li, Uwe Stilla, Joao~F Henriques, and Daniel Cremers.
\newblock Casspr: Cross attention single scan place recognition.
\newblock In \emph{ICCV}, pages 8461--8472, 2023.

\bibitem[Yang et~al.(2024)Yang, Li, Li, Cai, Wen, Zang, Muller, and Wang]{Yang_2024_CVPR}
Bochun Yang, Zijun Li, Wen Li, Zhipeng Cai, Chenglu Wen, Yu Zang, Matthias Muller, and Cheng Wang.
\newblock Lisa: Lidar localization with semantic awareness.
\newblock In \emph{CVPR}, pages 15271--15280, 2024.

\bibitem[Yin et~al.(2024)Yin, Xu, Lu, Chen, Xiong, Shen, Stachniss, and Wang]{Yin_2024_IJCV}
Huan Yin, Xuecheng Xu, Sha Lu, Xieyuanli Chen, Rong Xiong, Shaojie Shen, Cyrill Stachniss, and Yue Wang.
\newblock A survey on global lidar localization: Challenges, advances and open problems.
\newblock \emph{IJCV}, pages 1--33, 2024.

\bibitem[Yu et~al.(2021)Yu, Wang, Yu, Li, Cheng, and Zang]{Yu_2021_ISPRS}
Shangshu Yu, Cheng Wang, Zenglei Yu, Xin Li, Ming Cheng, and Yu Zang.
\newblock Deep regression for lidar-based localization in dense urban areas.
\newblock \emph{ISPRS-JPRS}, 172:\penalty0 240–252, 2021.

\bibitem[Yu et~al.(2022{\natexlab{a}})Yu, Wang, Lin, Wen, Cheng, and Hu]{Yu_2022_TITS}
Shangshu Yu, Cheng Wang, Yitai Lin, Chenglu Wen, Ming Cheng, and Guosheng Hu.
\newblock Stcloc: Deep lidar localization with spatio-temporal constraints.
\newblock \emph{IEEE TITS}, 24\penalty0 (1):\penalty0 489--500, 2022{\natexlab{a}}.

\bibitem[Yu et~al.(2022{\natexlab{b}})Yu, Wang, Wen, Cheng, Liu, Zhang, and Li]{Yu_2022_PR}
Shangshu Yu, Cheng Wang, Chenglu Wen, Ming Cheng, Minghao Liu, Zhihong Zhang, and Xin Li.
\newblock Lidar-based localization using universal encoding and memory-aware regression.
\newblock \emph{PR}, 128:\penalty0 108915, 2022{\natexlab{b}}.

\bibitem[Yu et~al.(2023)Yu, Sun, Li, Wen, Yang, Si, Hu, and Wang]{Yu_2023_TITS}
Shangshu Yu, Xiaotian Sun, Wen Li, Chenglu Wen, Yunuo Yang, Bailu Si, Guosheng Hu, and Cheng Wang.
\newblock Nidaloc: Neurobiologically inspired deep lidar localization.
\newblock \emph{IEEE TITS}, 2023.

\bibitem[Zhang et~al.(2014)Zhang, Singh, et~al.]{Zhang_2014_RSS}
Ji Zhang, Sanjiv Singh, et~al.
\newblock Loam: Lidar odometry and mapping in real-time.
\newblock In \emph{RSS}, pages 1--9, 2014.

\bibitem[Zhang et~al.(2024)Zhang, Wang, Chen, Zhang, and Hu]{Zhang_2024_AAAI}
Tongzhou Zhang, Gang Wang, Yu Chen, Hai Zhang, and Jue Hu.
\newblock Multi-constellation-inspired single-shot global lidar localization.
\newblock In \emph{AAAI}, pages 10404--10412, 2024.

\bibitem[Zhang et~al.(2023)Zhang, Yang, Zhang, and Zhang]{Zhang_2023_CVPR}
Xiyu Zhang, Jiaqi Yang, Shikun Zhang, and Yanning Zhang.
\newblock 3d registration with maximal cliques.
\newblock In \emph{CVPR}, pages 17745--17754, 2023.

\end{thebibliography}
}

\end{document}